\documentclass{article}
\usepackage{amsmath}
\usepackage{multirow}
\usepackage{PRIMEarxiv}
\usepackage{array}
\usepackage[utf8]{inputenc} % allow utf-8 input
\usepackage[T1]{fontenc}    % use 8-bit T1 fonts
\usepackage{hyperref}       % hyperlinks
\usepackage{url}            % simple URL typesetting
\usepackage{booktabs}       % professional-quality tables
\usepackage{amsfonts}       % blackboard math symbols
\usepackage{wrapfig}

\usepackage{nicefrac}       % compact symbols for 1/2, etc.
\usepackage{float}
\usepackage[dvipsnames]{xcolor}
\usepackage{tabularx}
\usepackage{microtype}      % microtypography
\usepackage{lipsum}
\usepackage{listings}
\usepackage{xcolor} % For color customization
\usepackage{fancyhdr}       % header
\usepackage{graphicx}       % graphics
\graphicspath{{media/}}     % organize your images and other figures under media/ folder
\title{ CEGI: measuring the trade-off between efficiency and  carbon emissions  for SLMs and VLMs}
\author{
  \begin{minipage}{0.5\textwidth}
    \centering
    \textbf{Abhas Kumar} \\
    Lead Data Scientist \\
    Synechron, Bangalore, India\\
    \texttt{abhas.kumar@synechron.com} 
  \end{minipage}%
  \hspace{0.001\textwidth}%
  \begin{minipage}{0.5\textwidth}
    \centering
    \textbf{Kapil Pathak} \\
    Lead Data Scientist \\
    Synechron, Bangalore, India\\
    \texttt{kapil.pathak@synechron.com}
  \end{minipage} \\[5em]  % Adjust this value to control row spacing
  \begin{minipage}{0.5\textwidth}
    \centering
    \textbf{Rajesh Kavuru} \\
    Lead Data Scientist \\
    Synechron, Bangalore, India\\
    \texttt{rajesh.kavuru@synechron.com} 
  \end{minipage}%
  \hspace{0.001\textwidth}%
  \begin{minipage}{0.5\textwidth}
    \centering
    \textbf{Prabhakar Srinivasan} \\
    Director  \\
    Synechron, Bangalore, India\\
    \texttt{prabhakar.srinivasan@synechron.com}
  \end{minipage}
}

\begin{document}
\maketitle

\lstset{
    language=Python,
    basicstyle=\ttfamily\small,
    keywordstyle=\color{blue}\bfseries,
    stringstyle=\color{purple}, % Default for single-line strings
    commentstyle=\color{green!70!black},
    breaklines=true,
    frame=single,
    numbers=left,
    numberstyle=\tiny,
    showstringspaces=false,
    moredelim=[s][\color{orange}\bfseries]{"""}{"""}, % Custom style for """<text>"""
}

% % keywords can be removed
% \keywords{First keyword \and Second keyword \and More}

\begin{abstract}
This paper analyzes the performance of Small Language Models (SLMs) and Vision Language Models (VLMs) and evaluates the trade-off between model performance and carbon emissions across 4 essential tasks: Image Captioning, Visual Question Answering (VQA), Dialogue
Summarization and Text-to-SQL conversion. Various SLMs and VLMs belonging to the Qwen and LLaMA architecture family are chosen and variants based on model size in terms of the number of parameters, quantization level and fine-tuning parameters are evaluated. The model variants' performance and carbon emissions are calculated. To quantify the trade-off between model performance and carbon emissions, we introduce a novel metric called CEGI (Carbon Efficient
Gain Index). This metric represents the carbon emission per unit percentage gain per million trainable parameters . This metric provides a normalized measure to compare models’ efficiency in terms
of performance improvement relative to their environmental cost. The experiment's outcome demonstrates
that fine-tuning SLMs and VLMs can achieve performance levels comparable to Large Language Models (LLMs) while
producing significantly less carbon emissions. Our findings suggest that the marginal gains in accuracy
from larger models do not justify the substantial increase in carbon emissions. Leveraging lower-bit
quantization levels, the proposed metric further enhances energy efficiency without compromising
performance. This study highlights balancing high performance and environmental sustainability. It offers a valuable metric for selecting models suitable for environmentally-friendly AI development.
\end{abstract}

% keywords can be removed
\keywords{Large Language Models (LLMs) \and Small Language Models (SLMs) \and Vision Language Models (VLMs) \and Carbon Emissions \and Sustainable AI \and Low-Rank Adaptation (LoRA) \and Quantization \and Green AI \and Model Fine-Tuning}

\section{Introduction}

The rapid advancements of Large Language Models (LLMs) and Vision Language Models (VLMs) have significantly advanced capabilities in complex tasks such as Visual Question Answering (VQA), Image Captioning, Dialogue Summarization, and Text-to-SQL Conversion. Models like Qwen and LLaMA have demonstrated remarkable performance in these domains. Huge computational requirements for pre-training and fine-tuning LLMs produces significant carbon emissions, raising environmental concerns. This study focuses on quantifying these emissions and assessing methods to mitigate them, addressing the need for efficient model adaptation that balances accuracy with reduced environmental impact.

Fine-tuning LLMs for domain-specific applications has become standard practice in Generative AI, allowing models to achieve high task-specific accuracy. However, recent studies highlight that this process consumes significant computational resources, leading to substantial carbon emissions. Strubell et al. (2019) \cite{strubell2019energypolicyconsiderationsdeep} and Patterson et al. (2021) \cite{patterson2021carbonemissionslargeneural} showed that training and fine-tuning modern models could emit as much \(\text{CO}_2\) as the lifetime emissions of multiple cars. We know that global carbon reduction goals are intensifying everywhere, and the AI community is responsible for addressing the environmental impact of model training and fine-tuning. This paper seeks to quantify and optimize these impacts by exploring efficient fine-tuning strategies to reduce emissions while maintaining performance.

While LLMs generally outperform SLMs in accuracy, the environmental impact of fine-tuning LLMs often outweighs marginal performance gains over SLMs. Techniques such as Low-Rank Adaptation (LoRA) \cite{hu2021loralowrankadaptationlarge} offer potential solutions by reducing the number of parameters updated during fine-tuning, enabling efficient domain adaptation in LLMs while preserving the baseline performance. With negligible loss in performance, lower-bit quantization techniques, such as 4-bit and 8-bit levels, can effectively decrease computational demands and energy consumption. These approaches show great promise in significantly reducing emissions while still preserving reliable task-level efficiency.
This study evaluates multiple model variants, focusing on Qwen and LLaMA architectures across 4 critical tasks: Image Captioning, Visual Question Answering, Dialogue Summarization, and Text-to-SQL Conversion. We analyze performance metrics and carbon emissions and compute requirements for each configuration using domain-specific datasets. The experiment leverages high-performance Graphical Processing Units (GPUs) for hardware acceleration, ensuring that results reflect real-world infrastructure. To quantify the trade-offs between performance and carbon emissions, we introduce a novel metric - the \textit{ unit percentage gain per million trainable parameters per gram of carbon emission}, i.e. CEGI. This metric provides a normalized measure to compare models' efficiency in terms of performance improvement relative to their environmental cost. Our work assesses the efficacy of fine-tuning smaller models and utilizing lower-bit quantization to mitigate model adaptation's trade-offs, thus informing sustainable AI practices.

Below are the major contributions of this paper:
\begin{itemize}
    \item \textbf{Carbon Footprint Quantification}: Empirical analysis of carbon emissions across various model configurations for key AI tasks, specifically focusing on Qwen and Llama architectures.
    \item \textbf{Novel Index}: Introduces a novel index, CEGI to quantify the trade-off between performance gain and environmental impact, facilitating more informed model selection.
    \item \textbf{Trade-Off Insights}: Comparative study of performance versus emissions in fine-tuning large models versus smaller models, highlighting that fine-tuned smaller models can achieve comparable performance with significantly lower carbon emissions.
    \item \textbf{Framework for Efficient Fine-Tuning}: Practical recommendations for adapting models sustainably using parameter-efficient fine-tuning methods like LoRA and lower-bit quantization without compromising accuracy.
\end{itemize}

\section{Related work}
\label{sec:headings}
The effectiveness of LLMs in a variety of Natural Language Processing (NLP) tasks has firmly established their position as pivotal assets in modern machine learning (Brown et al., 2020)\cite{brown2020languagemodelsfewshotlearners}. However, as these models scale up in complexity and are trained on increasingly larger datasets, the energy demands and environmental impact of deploying and operating them have intensified (Anil et al., 2023)\cite{anil2023palm2technicalreport}. With wide adoption of LLMs, integrated into many mainstream applications, their significant energy consumption has led to growing concerns about sustainability (Thompson et al., 2021)\cite{9563954}. For example, creating a 213-million-parameter transformer model through neural architecture search can produce ${CO}_2$ emissions equivalent to those generated by five cars over their entire lifespan (Strubell et al., 2019)\cite{strubell2019energypolicyconsiderationsdeep}.

While previous research has primarily concentrated on measuring the environmental impact of training large ML models, notably focusing on carbon emissions during this phase, it has tended to overlook the footprint associated with inference and fine-tuning. Research efforts, including those by Henderson et al. (2020)\cite{henderson2022systematicreportingenergycarbon}, Wu et al. (2022)\cite{wu2022sustainableaienvironmentalimplications}, Dodge et al. (2022)\cite{dodge2022measuringcarbonintensityai}, and Strubell et al. (2019)\cite{strubell2019energypolicyconsiderationsdeep}, have highlighted the substantial carbon emissions resulting from the training of large-scale models.

Our approach diverges by capturing carbon emissions across the entire lifecycle of model deployment, including training, fine-tuning, and inference. This study employs the eco2AI library (Budennyy et al., 2022)\cite{budennyy2022eco2aicarbonemissionstracking} to achieve detailed tracking of power consumptions and associated ${CO}_2$ emissions, ensuring accurate estimates across various hardware configurations and geographical emissions profiles. Unlike most of the other tools, eco2AI allows for nuanced tracking, accounting for hardware diversity and regional differences in emission factors, thus offering a precise measure of the environmental impact in terms of carbon emissions associated with LLM pretraining, fine-tuning, inference.

By evaluating carbon emissions across a range of downstream NLP tasks, like Visual-QA, Image Captioning, Text-To-SQL and Multiturn Dialogue Summarization, and implementing parameter-efficient strategies such as LoRA, this study aims to inform on sustainable AI practices. The findings aim to guide the development of computationally efficient architectures that maintain robust performance, advancing the broader objective of \textbf{Green AI}.

\section{Background}

\subsection{Eco2AI}
Eco2AI \cite{eco2AI} is an open-source library developed to help the data scientist measure the equivalent carbon emissions and power consumptions that occurred during the training or fine-tuning the deep neural networks. This library focuses on the calculation of electric energy consumption, extracting energy emission coefficients, and estimating the equivalent CO2 emissions. Eco2AI is able to detect GPU processes with the help of GPU management and monitoring functions implemented as a Python interface within the Pynvml library. The CPU utilization is monitored by $os$ and $psutil$ libraries. The emission intensities differ according to climate change, the region, the type of fuel used in that country, the economic and technological development of that region, etc. All these regional dependencies are accommodated by the emission intensity coefficient $\gamma$, which is the weight of $CO_{2}$ (in Kg) per each Mega Watt Hour (MWh) of the country. The final emission value as a carbon footprint of the ML processes is calculated by the following formula:
\begin{equation}
  CF = \gamma*PUE*(E_{CPU}+E_{GPU}+E_{RAM})
\end{equation}
where CF is a carbon footprint of the ML process, $PUE$ is
the power usage effectiveness of the data center if any kind of cloud is used in the process. $E_{CPU}$, $E_{GPU}$, and $E_{RAM}$ are the power consumption values from CPU, GPU, and RAM, respectively. 

\subsection{Low Rank Adaptation (LoRA)}
Most of the applications in NLP rely on adapting the pre-trained model, which is trained on large-scale data and on specific tasks with limited task-specific data. The resulting fine-tuned model also consists of the same number of trainable parameters as the original pre-trained model. This kind of operational inefficiency can be handled by adding a small number of trainable parameters to the non-trainable weights of the original pre-trained model. In these efforts, Edward et al. \cite{hu2021loralowrankadaptationlarge} proposed a novel method Low-Rank Adaptation (LoRA) which is inspired by the hypothesis \cite{li2018measuringintrinsicdimensionobjective}, \cite{aghajanyan-etal-2021-intrinsic} that the trained over-parametrized models actually can be explained by the lower intrinsic dimension. They propose that the weight changes in the fine-tuning process also reside in the low intrinsic space. LoRA allows us to train only a small number of dense layers by optimizing the rank-decomposition matrices of the trainable dense layers and keeps most of the weights of the pre-trained model frozen.

For a pre-trained weight matrix $W_{0} \in \mathbb{R}^{d \times k}$, the weight updates are represented as $\Delta W$, which can be decomposed into low-rank matrices $A \in \mathbb{R}^{r \times k}$ and $B \in \mathbb{R}^{d \times r}$, the rank $r \ll min(d, k)$. 
\begin{equation}
  W_{0} + \Delta W = W_{0} + BA
\end{equation}
By this approach, we can share the large pre-trained model among different tasks and fine-tune many task-specific small LoRA modules. LoRA makes fine-tuning more hardware efficient as we don't need to calculate the gradients and store the optimizer states for all parameters of the pre-trained model. Here, both matrices $W_{O}$ and $\Delta W$ are multiplied by the same input. The corresponding output representations are summed up, and the resulting forward pass for input $x$ can be given below. 
\begin{equation}
  h = W_{O}x + \Delta Wx = W_{0}x + BAx
\end{equation}
\subsection{Visual Question Answering}
Visual Question Answering (VQA) is the task of answering open-ended questions based on the given image. In this task, we give an image and a question as input to the model and get an answer in the form of text as an output. In this task, the models are expected to understand the content of the image and the semantics of the questions. The interaction between natural language and the image content should enable the model to respond to the questions effectively. TY Lin et al. \cite{mscoco} presented the dataset MS-COCO for various image understanding tasks. In their work, Zhan et al. \cite{9999450} proposed a novel conditional reasoning method for the medical visual question-answering task. Biten et al. \cite{biten2019scenetextvisualquestion} presented a new dataset ST-VQA, which aims at exploiting the high-level semantic information given in the images through textual cues in the VQA task. 
\subsection{Image Captioning}
Image Captioning is another task that involves an image as an input but expects the natural language as an output. In this task, the model aims at describing the content of the image. MS COCO \cite{mscoco} is one of the first notable image captioning datasets in this domain. The image captioning models are based on an encoding-decoding framework where the image as input is encoded into the intermediate representation by the encoder and decoded by the decoder into the descriptive natural language. The modern image captioning architectures are transformer-based \cite{NIPS2017_3f5ee243} while improving encoder-decoder formulation \cite{9157506}, \cite{Herdade2019ImageCT}, \cite{guo-etal-2019-star}. The Visual Language Models (VLMs) are inspired by contrastive-based architectures such as CLIP \cite{radford2021learningtransferablevisualmodels} where the positive data points are sampled from the ground truth image-caption labels and the negative data points are sampled from other captions of the same mini-batch. The novel aspect of the CLIP model is that the image representation and the text representation are brought in the same space. 
\subsection{Dialogue Summarization}
The abstractive dialogue summarization is a text-to-text generation task where the input is a multi-turn conversation and generates its summary. There are various benchmark datasets such as SAMSum \cite{gliwa-etal-2019-samsum}, DialogSum \cite{chen-etal-2021-dialogsum}, MediaSum \cite{zhu-etal-2021-mediasum} etc. The SAMSum \cite{gliwa-etal-2019-samsum} dataset consists of short conversations of leisure chitchat. DialogSum dataset \cite{chen-etal-2021-dialogsum} comprises the conversations in the real-world scenarios. MediumSum \cite{zhu-etal-2021-mediasum} consists of large-scale media interviews with 463.6K transcripts. Traditionally, text-to-text tasks such as dialogue summarization follow the encoder-decoder architecture, where the input text is encoded into a context vector and decoded by the decoder in the form of output text. 
\subsection{Text-to-SQL Generation}
Text-to-SQL generation is a task in natural language processing whose aim is to generate SQL queries based on the natural language. Yu et al. \cite{yu-etal-2018-spider} presented Spider, a large-scale complex and cross-domain text-to-SQL dataset that became one of the first benchmarks for text-to-SQL tasks. Lee et al. \cite{lee-etal-2021-kaggledbqa} introduced KaggleDBQA, a domain evaluation dataset of real Web databases. This dataset is also featured with domain-specific data types, original formatting, and unrestricted questions.

\section{Methodology}

This section mainly outlines the experimental framework we employed to investigate the trade-offs between carbon emissions and model performance in fine-tuning large and small language models across four downstream NLP tasks. The methodology includes details of the datasets, models, fine-tuning techniques, and carbon emission tracking tools used. Each component is carefully designed to ensure the reproducibility and rigor of the experiments.

\subsection{Datasets and Tasks}
Our study focuses on four tasks: \textbf{Image Captioning}, \textbf{Visual Question Answering (VQA)}, \textbf{Dialogue Summarization}, and \textbf{Text-to-SQL}, selected for their diversity in modality and complexity to serve text-to-text and image-to-tasks. The data sets used for each task are curated to reflect real-world domain adaptation scenarios, ensuring the robustness of the analysis. Appendix section \ref{sec:dataset} provides additional information on all datasets used.

For the Image Captioning, we used the \textbf{ artbench-pd-256x256}\cite{liao2022artbench} data set, a subset of ArtBench. ArtBench\footnote{\href{https://huggingface.co/datasets/alfredplpl/artbench-pd-256x256}{Huggingface alfredplpl/artbench-pd-256x256 }.} is the dataset for historical arts such as Art Nouveau and Ukiyo-e picked from public domain images from ArtBench. We have used 2,000 training and 200 test samples, with evaluation based on SPICE scores, to capture semantic accuracy in generated captions. For Visual Question Answering (VQA), \textbf{PathVQA}\cite{he2020pathvqa} is a data set designed to train and evaluate visual medical question-answer systems (VQA) using pathology images\footnote{\href{https://huggingface.co/datasets/flaviagiammarino/path-vqa}{Huggingface flaviagiammarino/path-vqa}.}, is tailored to image-question pairs, evaluated using BLEU(Bilingual Evaluation Understudy)\cite{papineni2002bleu} scores to measure language-based precision.

For the Dialogue Summarization, \textbf{SAMSum}\cite{gliwa-etal-2019-samsum} dataset has been used here, which comprises of messenger-style conversations\footnote{\href{https://huggingface.co/datasets/gretelai/synthetic_text_to_sql}{Huggingface knkarthick/samsum}} The data set includes 14,000 training samples and 1,400 test samples, with performance evaluated using Recall-Oriented Understudy for Gisting Evaluation metric (ROUGE-1,2,L)\cite{lin-2004-rouge} metrics. For text-to-SQL task, \textbf{synthetic\_text\_to\_sql}\cite{gretel-synthetic-text-to-sql-2024}This data set targets SQL query generation\footnote{\href{https://huggingface.co/datasets/gretelai/synthetic_text_to_sql}{gretelai/synthetic\_text\_to\_sql}} tasks, evaluated using Execution Accuracy (EA) and Valid Efficiency Score (VES) metrics. The diversity in tasks and evaluation metrics ensures that the analysis captures nuanced trade-offs between emissions and performance across distinct tasks.

\begin{table}[ht]
    \centering
    \begin{tabular}{|p{0.15\textwidth}|p{0.40\textwidth}|p{0.08\textwidth}|p{0.08\textwidth}|p{0.15\textwidth}|}
        \hline
        \textbf{Task} & \textbf{Models($M$)} & \textbf{Train Size} & \textbf{Test Size} & \textbf{ Metric} \\ 
        \hline
        Summarization & \multirow{2}{=}{Qwen2.5-0.5B, Qwen2.5-3B, Qwen2.5-7B, Qwen2.5-14B, Llama-3.2-1B, Llama-3.2-3B} & 14000 & 1400 & ROUGE(1,2,L) \\ 
        \cline{1-1} \cline{5-5}
        Test-to-SQL & & & & EA, VES \\ 
        \hline
        Image Captioning & \multirow{2}{=}{Qwen-VL-2B, Qwen-VL-7B, Llama-3.2-11B} & 2000 & 200 & SPICE \\ 
        \cline{1-1} \cline{5-5}
        Visual QA & & & & BLUE \\ 
        \hline
    \end{tabular}
    \vspace{2mm}
    \caption{Summary of tasks, models, train/test sizes, and evaluation metrics}
    \label{tab:task_summary}
\end{table}

\subsection{Model Configurations}
We utilize diverse models for carbon tracking experiments to investigate the relationship between model size, performance, and emissions across tasks. For Dialogue Summarization and Text-to-SQL, we employ the \textbf{Qwen2.5} instruct model family (0.5B, 3B, 7B, 14B) and \textbf{LLaMA3.2} instruct (1B, 3B). For Image-to-Text tasks, namely Image Captioning and Visual-QA, we used Qwen-VL instruct (2B, 7B) and LLaMA-3.2-11B Vision-Instruct  LLMs. 

These models, as summarized in Table \ref{tab:task_summary}, span a range of parameter scales, from smaller, efficiency-focused architectures to large-scale models designed for high-capacity tasks. This diversity allows us to evaluate the carbon emissions generated during fine-tuning and inference systematically. Baseline evaluations were conducted to establish each model's emissions and performance before fine-tuning, enabling direct comparisons with parameter-efficient methods such as LoRA and quantization.

\subsection{Fine-Tuning Technique}

We have used Low-Rank Adaptation (LoRA) to fine-tune models efficiently. As known LoRA introduces low-rank matrices into existing layers, enabling domain adaptation with minimal parameter updates. This method reduces the computational cost associated with full fine-tuning while maintaining task-specific accuracy.

To evaluate the trade-offs between performance and resource usage, we experiment with four LoRA rank configurations ($\mathbf{L_r=4, 8, 16, 32}$). Additionally, quantization \boldmath{$Q_b$ } is applied to compress model parameters to 4-bit and 8-bit precision, further reducing computational overhead and energy consumption. These configurations enable a detailed analysis of parameter-efficient fine-tuning across various scenarios.

\subsection{Hardware Configurations and Carbon Emissions Tracking}
All experiments are conducted on \textbf{NVIDIA 1xH100(80 GB)} GPU, selected for its high computational throughput and energy efficiency. To measure the environmental impact, we utilize the \textbf{Eco2AI} library (Budennyy et al., 2022)\cite{budennyy2022eco2aicarbonemissionstracking}, an open-source tool for p.recise energy consumption and carbon emission tracking. Eco2AI integrates the following:
\begin{itemize}
    \item Hardware Metrics: Tracks GPU utilization, power consumption, and training duration.
    \item Regional Emissions Factors: Incorporates electricity grid carbon intensity variations, ensuring context-aware emission estimates.
This approach enables granular emissions measurement for fine-tuning and inference, providing robust data for analyzing configuration trade-offs.
\end{itemize}

\subsection{Evaluation Metrics}
Performance is assessed using task-specific metrics to ensure relevance to each domain:
\begin{itemize}
    \item \textbf{Image Captioning}: \textbf{SPICE}\cite{DBLP:journals/corr/AndersonFJG16} scores, which evaluate the semantic relevance of generated captions.
    \item \textbf{Visual Question Answering (VQA)}: \textbf{Smoothed BLEU} scores, capturing linguistic precision and fluency.
    \item \textbf{Dialogue Summarization}: \textbf{ROUGE} scores (ROUGE-1, ROUGE-2, ROUGE-L), measuring n-gram overlaps with reference summaries.
    \item \textbf{Text-to-SQL}: \textbf{Execution Accuracy (EA)} and \textbf{Valid Efficiency Score (VES)}, assessing the correctness and semantic equivalence of SQL queries.
\end{itemize}

\textbf{SPICE} evaluates semantic content by analyzing the alignment of objects, attributes, and relationships between reference and generated outputs, prioritizing semantic consistency over lexical similarity. Smoothed BLEU refines the traditional \textbf{BLEU} metric by addressing its limitations in handling exact n-gram matches and rare patterns. Smoothing techniques enhance reliability for evaluating shorter sequences and partial matches in text generation tasks. \textbf{EA (Execution Accuracy)} quantifies the correctness of generated SQL queries by comparing them to ground-truth outputs, with scores ranging from 0 to 1, where 1 indicates complete accuracy. \textbf{VES (Valid Efficiency Score)} assesses the efficiency of generated SQL queries relative to reference queries, with values approaching 1 signifying equivalent or superior query execution performance.

\section{Experiments}
For all tasks, fine-tuning was performed for $\mathbf{1}$ epoch to ensure uniformity across experiments. Each reported value, including performance metrics and emissions, represents the mean of $\mathbf{5}$ independent runs, ensuring stability and reliability in the results. Additionally, we used \textbf{GPT-4o} in a zero-shot setup as a baseline for comparative evaluation. Carbon emissions data for GPT-4o cannot be obtained since it is a proprietary model. The detailed prompts used for Qwen2.5, Qwen-VL LLaMa3.2, and GPT-4o, as well as fine-tuning configurations for each task, are provided in the prompt section\ref{sec:prompts} of the appendix for reproducibility. Note that the Base model(pre-trained) and fine-tuned models are represented as $\mathbf{B_{M}}$ and $\mathbf{F_{T}}$, carbon emissions \boldmath{$C_E$} in grams. 

The LoRA rank was varied across ($\mathbf{L_r=4, 8,16,32}$), allowing us to systematically study the trade-offs between the number of updated parameters and the resulting task performance with corresponding $CO_2$ emissions. A dropout rate of 5 percent is applied to introduce regularization during fine-tuning stage, mitigating the potential for overfitting. The target modules for LoRA focused on critical components of transformer architectures, including projection layers $\mathbf{([Q, K, V, O])_{proj}}$ and gating mechanisms $\mathbf{([gate, up, down])_{proj}}$. These layers are fundamental to the transformer’s ability to model complex relationships in data, and fine-tuning them ensures efficient domain adaptation with causal language modeling tasks. The scaling parameter $\boldsymbol{\alpha}$ was set to $\mathbf{16}$, controlling the magnitude of low-rank updates and ensuring effective adaptation without destabilizing the pre-trained parameters of the base model.

LLM \textit{temperature} set to $\mathbf{0}$ for fully deterministic outputs, ensuring consistent and predictable results without any randomness and  \textit{Maximum Token Length} capped at $\mathbf{1024}$ to ensure inference efficiency without truncation of outputs.

\subsection{Image Captioning and Visual-QA}

For Image Captioning(Table \ref{tab:spice_scores}) and Visual-QA(Table \ref{tab:visual_qa_scores}), we conducted experiments using Qwen-VL models (2B and 7B) and LLaMA-3.2-11B Vision-Instruct models as the primary architectures. In both tasks, the models were fine-tuned using Low-Rank Adaptation (LoRA) with ranks ($L_r=8,16,32$) and 4-bit quantization ($Q_b=4$), enabling efficient parameter adaptation while minimizing computational overhead. The evaluation metrics used were SPICE (Semantic Propositional Image Caption Evaluation) for Image Captioning and BLEU for Visual-QA, which measure semantic relevance and language-based precision, respectively. GPT-4o under a zero-shot setup was employed as a baseline for both tasks to benchmark the performance of fine-tuned models. Prompts tailored to Image Captioning and Visual-QA tasks (\ref{sec:prompts}) were applied uniformly across all models to ensure consistency. Carbon emissions ($C_E$) were recorded for each configuration, providing insights into the environmental cost of fine-tuning. 

GPT-4o served as a reference point, offering competitive zero-shot performance. Still, fine-tuned models surpassed GPT-4o with higher SPICE and BLEU scores, demonstrating that specialized models can outperform general-purpose LLMs. Overall, these experiments highlight the trade-off between performance improvements and environmental impact, with smaller fine-tuned models like Qwen-VL-2B and Qwen-VL-7B achieving a favorable balance between performance and minimal carbon emissions.

\begin{table}[h]
    \centering
    \begin{tabular}{l|l|c|c|c|c|c}
        \toprule
        \textbf{Model($M$)} & \boldmath{$L_r$} & \boldmath{$Q_b$} & \boldmath{$B_{M}$} \textbf{SPICE} & \boldmath{$F_{T}$} \textbf{SPICE} & \boldmath{$C_{E}$(gm)} & \textbf{GPT-4o SPICE} \\
        \midrule
        \multirow{3}{*}{Qwen-VL-2B} & 8  & \multirow{9}{*}{4-bit} &       & 0.35 & 95.5 (\textbf{base})  & \multirow{9}{*}{0.16} \\
                                     & 16 &                    & 0.16  & 0.36 & 97.8  (+2.4\%) &                       \\
                                     & 32 &                    &       & 0.35 & 98.8  (+3.4\%) &                       \\
        \cmidrule(lr){1-2} \cmidrule(lr){4-6}
        \multirow{3}{*}{Qwen-VL-7B}  & 8  &                    &       & 0.37 & 137.8 (+44.3\%) &                       \\
                                     & 16 &                    & 0.14  & 0.37 & 137.9 (+44.3\%) &                       \\
                                     & 32 &                    &       & 0.38 & 138.2  (+44.7\%)&                       \\
        \cmidrule(lr){1-2} \cmidrule(lr){4-6}
        \multirow{3}{*}{Llama-3.2-11B} & 8 &                   &       & 0.30 & 222.11 (+132.58\%)&                       \\
                                       & 16 &                  & 0.18  & 0.29 & 222.46 (+\textbf{133.\%})&                       \\
                                       & 32 &                  &       & 0.31 & 221.51 (132\%)&                       \\ 
        \bottomrule
    \end{tabular}
    \vspace{2mm}
    \caption{Comparison of 4-bit base and fine-tuned model variants, ranks, SPICE scores, and carbon emissions for image captioning tasks.}
    \label{tab:spice_scores}
\end{table}

\begin{table}[H]
    \centering
    \begin{tabular}{l|l|c|c|c|c|c}
        \toprule
        \textbf{Model($M$)} & \boldmath{$L_r$} & \boldmath{$Q_b$} & \boldmath{$B_{M}$}\textbf{BLEU} & \boldmath{$F_{T}$} \textbf{BLEU} & \boldmath{$C_{E}$(gm)} & \textbf{GPT-4o BLEU} \\
        \midrule
        \multirow{3}{*}{Qwen-VL-2B} & 8  & \multirow{9}{*}{4-bit} & 0.0037 & 0.0552 & 60.1  (\textbf{base}) & \multirow{9}{*}{0.0013} \\
                                     & 16 &                    &        & 0.0592 & 60.7  (+1\%) &                       \\
                                     & 32 &                    &        & 0.0587 & 60.5  (+0.6\%) &                       \\
        \cmidrule(lr){1-2} \cmidrule(lr){4-6}
        \multirow{3}{*}{Qwen-VL-7B}  & 8  &                    & 0.0010  & 0.0732 & 78.8 (+31.1\%) &                       \\
                                     & 16 &                    &        & 0.0767 & 77.3 (+28.6\%) &                       \\
                                     & 32 &                    &        & 0.075  & 78.9  (+31.3\%) &                       \\
        \cmidrule(lr){1-2} \cmidrule(lr){4-6}
        \multirow{3}{*}{Llama-3.2-11B} & 8 &                   & 0.0006 & 0.0014 & 212.82 (+254.1\%)&                       \\
                                       & 16 &                  &        & 0.0017 & 211.57 (+252.03\%)&                       \\
                                       & 32 &                  &        & 0.0017 & 213.9 (+\textbf{256.0\%}) &                       \\
        \bottomrule
    \end{tabular}
    \vspace{2mm}
    \caption{Comparison of 4-bit base and fine-tuned model model variants, ranks, BLEU scores, and carbon emissions for visual-QA task.}
    \label{tab:visual_qa_scores}
\end{table}

\subsection{Summarization and Text-to-SQL}

We fine-tuned Qwen2.5 models (0.5B, 3B, 7B, 14B) and LLaMA models (1B, 3B) for Dialogue Summarization and Text-to-SQL tasks. These models were fine-tuned using Low-Rank Adaptation (LoRA) with ranks (${L_r=4, 8, 16}$) and quantization levels ($Q_b = 4,8$). This systematic approach enabled efficient adaptation of the models to the specific requirements of each task while minimizing computational demands. For Dialogue Summarization, ROUGE scores (ROUGE-1, ROUGE-2, and ROUGE-L) were employed as evaluation metrics to measure the quality and relevance of generated summaries ( Table \ref{tab:summarization_scores}).

For the Text-to-SQL task, Execution Accuracy (EA) and Valid Efficiency Score (VES) were used as the primary evaluation metrics (Table \ref{tab:text_to_sql_scores}). These metrics assessed the correctness of generated SQL queries and their execution validity across datasets. Dialogue summarization-specific and SQL-specific prompts (\ref{sec:prompts}) were used during the fine-tuning process and for GPT-4o inference is under a zero-shot setup. The prompts are tailored to ensure consistency and optimal performance across all model configurations. Complete experimental results, including detailed performance and carbon emissions comparisons for both tasks, are summarized in (\ref{sec:experiments}).

\begin{table}[H]
    \centering
    \begin{tabular}{l|c|c|c|c|c|c}
        \toprule
        \textbf{Model ($M$)} & \boldmath{$L_r$} & \boldmath{$Q_b$} & \boldmath{$B_{M}$} \textbf{ROUGE-1} & \boldmath{$F_{T}$} \textbf{ROUGE-1} & \boldmath{$C_E$}\textbf{ (gm)} & \textbf{GPT-4o} \\
        \midrule
        % Qwen2.5-0.5B
        \multirow{3}{*}{Qwen2.5-0.5B} 
        & 4  & \multirow{3}{*}{} & 0.23 & 0.42 & 25.2 (\textbf{base}) & \multirow{18}{*}{} \\
        & 8  &                        & 0.23 & 0.43 & 25.3 (+0.4\%) &  \\
        & 16 &                        & 0.23 & 0.43 & 25.4 (+0.8\%) &  \\
        \cmidrule(lr){1-2} \cmidrule(lr){4-6}
        
        % Qwen2.5-3B
        \multirow{3}{*}{Qwen2.5-3B} 
        & 4  & \multirow{3}{*}{} & 0.30 & 0.50 & 54.3 (+115.5\%) &  \\
        & 8  &                        & 0.30 & 0.49 & 55.2 (+119.0\%) &  \\
        & 16 &                        & 0.30 & 0.49 & 54.3 (+115.5\%) &  \\
        \cmidrule(lr){1-2} \cmidrule(lr){4-6}
        
        % Qwen2.5-7B
        \multirow{3}{*}{Qwen2.5-7B} 
        & 4  & \multirow{3}{*}{} & 0.32 & 0.50 & 77.2 (+206.3\%) &  \\
        & 8  &                        & 0.32 & 0.50 & 77.3 (+206.7\%) &  \\
        & 16 &                 8-bit       & 0.32 & 0.50 & 77.3 (+206.7\%) & 0.37 \\
        \cmidrule(lr){1-2} \cmidrule(lr){4-6}
        
        % Qwen2.5-14B
        \multirow{3}{*}{Qwen2.5-14B} 
        & 4  & \multirow{3}{*}{} & 0.33 & 0.51 & 154.6 (+513.5\%) &  \\
        & 8  &                        & 0.33 & 0.52 & 152.5 (+505.6\%) &  \\
        & 16 &                        & 0.33 & 0.51 & 158.4 (+\textbf{528.6\%}) &  \\
        \cmidrule(lr){1-2} \cmidrule(lr){4-6}
        
        % Llama-3.2-1B
        \multirow{3}{*}{Llama-3.2-1B} 
        & 4  & \multirow{3}{*}{} & 0.26 & 0.49 & 37.6 (+49.2\%) &  \\
        & 8  &                        & 0.26 & 0.49 & 39.2 (+55.6\%) &  \\
        & 16 &                        & 0.26 & 0.49 & 36.4 (+44.4\%) &  \\
        \cmidrule(lr){1-2} \cmidrule(lr){4-6}
        
        % Llama-3.2-3B
        \multirow{3}{*}{Llama-3.2-3B} 
        & 4  & \multirow{3}{*}{} & 0.28 & 0.50 & 70.3 (+179.0\%) &  \\
        & 8  &                        & 0.28 & 0.47 & 70.2 (+178.6\%) &  \\
        & 16 &                        & 0.28 & 0.48 & 68.6 (+172.2\%) &  \\
        \bottomrule
    \end{tabular}
    \vspace{2mm}
    \caption{Comparison of 8-bit base and fine-tuned model variants, LoRA ranks, ROUGE-1 scores, and carbon emissions (with percentage increase from the base value) for the dialogue summarization task.}
    \label{tab:summarization_scores}
\end{table}

\begin{table}[H]
    \centering
    \begin{tabular}{l|l|c|c|c|c|c}
        \toprule
        \textbf{Model ($M$)} & \boldmath{$L_r$} & \boldmath{$Q_b$} &  \boldmath{$(B_{M},F_{T})$ EA} & \boldmath{$(B_{M},F_{T})$ VES} & \boldmath{$C_E$ (g)} & \textbf{GPT-4o (EA, VES)} \\
        \midrule
        % Qwen2.5-0.5B
        \multirow{3}{*}{Qwen2.5-0.5B} 
            & 4  & \multirow{3}{*}{} & 0.42, 0.72 & 0.97, 0.98 & 30.71 (\textbf{base}) & \multirow{18}{*}{0.80, 0.98} \\
            & 8  &                         & 0.42, 0.73 &  0.97, 0.98 & 31.16 (+1.5\%) &  \\
            & 16 &                         & 0.42, 0.72 & 0.97, 0.98 & 31.06 (+1.1\%) &  \\
        \cmidrule(lr){1-2} \cmidrule(lr){4-6}
        % Qwen2.5-3B
        \multirow{3}{*}{Qwen2.5-3B} 
            & 4  & \multirow{3}{*}{} & 0.43, 0.79 & 0.97, 0.99 &   77.69 (+153.0\%) &\\
            & 8  &                         & 0.43, 0.79 & 0.97, 0.98 & 77.23 (+151.5\%) &  \\
            & 16 &                         & 0.43, 0.79 & 0.97, 0.98 & 73.75 (+140.2\%) &  \\
        \cmidrule(lr){1-2} \cmidrule(lr){4-6}
        % Qwen2.5-7B
        \multirow{3}{*}{Qwen2.5-7B} 
            & 4  & \multirow{3}{*}{} & 0.45, 0.78 & 0.98, 0.99 & 122.86 (+300.1\%) &  \\
            & 8  &                         & 0.45, 0.77 & 0.98, 0.98 & 122.85 (+300.1\%) &  \\
            & 16 &               8          & 0.45, 0.77 & 0.98, 0.98 & 122.09 (+297.5\%) &  \\
        \cmidrule(lr){1-2} \cmidrule(lr){4-6}
        % Qwen2.5-14B
        \multirow{3}{*}{Qwen2.5-14B} 
            & 4  & \multirow{3}{*}{} & 0.46, 0.82 & 0.98, 0.99 &  236.69 (+\textbf{671.0\%})& \\
            & 8  &                         & 0.46, 0.81 & 0.98, 0.99 & 235.97 (+668.5\%) &  \\
            & 16 &                         &  0.46, 0.81 & 0.98, 0.99 & 234.85 (+664.9\%) & \\
        \cmidrule(lr){1-2} \cmidrule(lr){4-6}
        % Llama-3.2-1B
        \multirow{3}{*}{Llama-3.2-1B} 
            & 4  & \multirow{3}{*}{} & 0.45, 0.75 & 0.97, 0.98 &  52.69 (+71.6\%) & \\
            & 8  & & 0.45, 0.75  & 0.97, 0.98 & 51.49 (+67.7\%) &  \\
            & 16 &                         &  0.45, 0.75 & 0.97, 0.98 & 51.86 (+68.9\%) & \\
        \cmidrule(lr){1-2} \cmidrule(lr){4-6}
        % Llama-3.2-3B
        \multirow{3}{*}{Llama-3.2-3B} 
            & 4  & \multirow{3}{*}{} & 0.42, 0.77 & 0.98, 0.99 & 104.36 (+239.8\%) &  \\
            & 8  &                         & 0.42, 0.78 & 0.98, 0.98 & 94.42 (+207.4\%) &  \\
            & 16 &                         & 0.42, 0.79 & 0.98, 0.98 &  94.48 (+207.6\%) & \\
        \bottomrule
    \end{tabular}
    \vspace{2mm}
    \caption{Comparison of 8-bit base and fine-tuned model variants, LoRA ranks, EA, VES scores, and carbon emissions (with percentage increase from the base value) for the text-to-SQL task.}
    \label{tab:text_to_sql_scores}
\end{table}

\section{Results}
This section provides a comprehensive quantitative analysis of the significant performance improvements achieved through fine-tuning models across four key tasks: Image Captioning, Visual Question Answering (VQA), Dialogue Summarization, and Text-to-SQL conversion. These insights highlight the potential of fine-tuning in enhancing AI model performance. This analysis examines the performance improvements achieved through fine-tuning compared to base models and benchmarks these results against GPT-4o under a zero-shot setting for each task. Additionally, we explore the trade-offs among model size, performance, and carbon emissions, highlighting their interconnection in promoting sustainable AI practices. Our findings reveal that fine-tuning smaller models can achieve competitive performance while significantly reducing carbon emissions, offering a viable alternative to deploying larger, resource-intensive models. By incorporating lower-bit quantization levels (e.g., 4-bit and 8-bit), we demonstrate the feasibility of maintaining energy efficiency without compromising task-specific model performance. The results underscore the importance of optimizing fine-tuning strategies to balance environmental responsibility and the effective development of high-performance AI models which are aligned with sustainable AI practices.

\subsection{Performance Gain}

We have measured the performance gain of a fine-tuned model compared to the base model and GPT-4o (under zero-shot setup) relevant across tasks, the gain for a particular metric, $\mu$ that is,$\text{G}_{M,\mu}$ is calculated as the percentage improvement of the metric value from a reference model to a target model using

\begin{equation}
    \text{G}_{M,\mu} (\%) = \left( \frac{\mu_{T=\{F_T\}} - \mu_{R=\{B_M, \text{GPT-4o}\}}}{\mu_{R=\{B_M, \text{GPT-4o}\}}} \right) \times 100\%
\end{equation}

\noindent where, 

\begin{itemize}
    \item $\mu_T$: Metric value of the Target Model e.g., fine-tuned \( F_T \) model. 
    \item $\mu_R$: Metric value of the Reference Model, which can be  either (base model) \( B_M \)  or \( \text{GPT-4o} \), depending on the comparison.
    \item \textbf{G\(_{M,\mu}\) (\%)}: Percentage gain for a specific metric \( \mu \) (e.g., ROUGE, BLEU, SPICE, EA, VES) from the reference model to the target model.
\end{itemize}

\subsubsection{Image Captioning and visual-QA}

\textbf{Image Captioning:} GPT-4o has a SPICE score(Table \ref{tab:spice_scores}) of $0.16$ for image captioning task on the artbench-pd-256x256 dataset\cite{liao2022artbench} matching the base Qwen-VL-2B but significantly lower than corresponding fine-tuned models.
\begin{figure}[H]
    \centering
    \begin{minipage}[t]{0.48\linewidth}
        \centering
        \includegraphics[width=\linewidth]{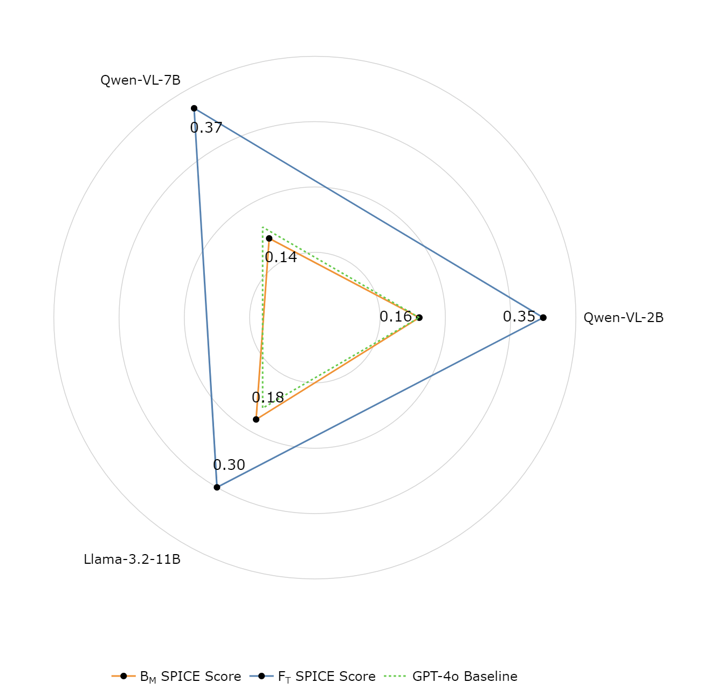}
        \caption{Performance comparison for image captioning tasks using SPICE score. The plot illustrates the performance of the base model ($B_M$), fine-tuned model ($F_T$), and GPT-4o as the baseline. Fine-tuned models demonstrate superior SPICE scores across configurations.}
        \label{fig:figure_image_caption_performance}
    \end{minipage}
    \hfill
    \begin{minipage}[t]{0.48\linewidth}
        \centering
        \includegraphics[width=\linewidth]{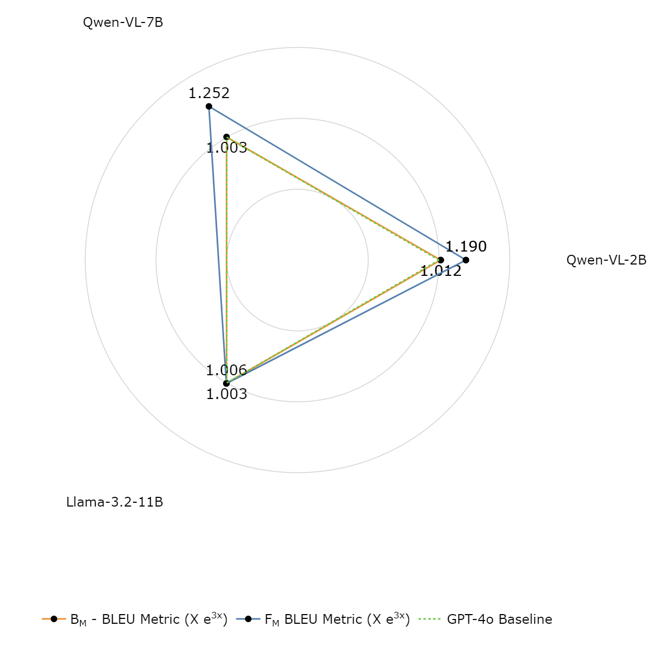}
        \caption{Performance comparison(on $e^{3x}$ scale) for Visual QA tasks using BLEU score. The radar chart highlights the improvements achieved by the fine-tuned model ($F_T$) compared to the base model ($B_M$) and GPT-4o baseline, demonstrating substantial performance gains in fine-tuned configurations.}
        \label{fig:figure_visual_qa_performance}
    \end{minipage}
\end{figure}

\begin{table}[H]
    \centering
    \begin{tabular}{l|c|c|c|c|c}
        \toprule
        \textbf{Model($M$)} & \boldmath{$L_r$} & \boldmath{$B_M$} \textbf{SPICE} & \boldmath{$F_T$} \textbf{SPICE} & \textbf{Gain(\boldmath{$F_T$} VS \boldmath{$B_M$})} & \textbf{Gain(\boldmath{$F_T$} VS \textbf{$GPT-4o$})}   \\ 
        \midrule
        \multirow{3}{*}{Qwen-VL-2B} 
            & 8  &  & 0.35 & +118.75 \% & +118.75 \%  \\
            & 16 & 0.16 & 0.36 & +125.00 \% & +125.00 \% \\
            & 32 &  & 0.35 & +118.75 \% & +118.75 \%  \\
        \cmidrule(lr){1-6}
        \multirow{3}{*}{Qwen-VL-7B} 
            & 8  &  & 0.37 & +164.29 \% & +131.25 \% \\
            & 16 & 0.14 & 0.37 & +164.29 \% & +131.25 \%   \\
            & 32 &  & \textbf{0.38} & \textbf{+171.43} \% & \textbf{+137.50} \%   \\
        \cmidrule(lr){1-6}
        \multirow{3}{*}{Llama-3.2-11B} 
            & 8  &  & 0.30 & +66.67 \%  & +87.50 \% \\
            & 16 & \textbf{0.18} & 0.29 & +61.11 \%  & +81.25 \%  \\
            & 32 &  & 0.31 & +72.22 \%  & +93.75 \%  \\
        \bottomrule
    \end{tabular}
    \vspace{2mm}
    \caption{Performance gain after Fine-tuning compared to $B_M$ and GPT-4o for image captioning task}
    \label{tab:spice_improvement}
\end{table}

After Fine-tuning, Qwen-VL-7B achieves 171.43 \% gain and also achieves the highest performance gain(chart \ref{fig:figure_image_caption_performance}) of 137.50 \% compared to GPT-4o. The larger model, like Qwen-VL-7B, compared to Qwen-VL-2B, provides higher absolute SPICE scores. However, the percentage improvement is more significant in smaller models (e.g., Qwen-VL-7B's 171.43 \% vs. Llama-3.2-11B's 72.22\%). Qwen-VL-2B emits 95.5g, whereas Llama-3.2-11B emits 222.11g(Table \ref{tab:spice_scores}). The marginal SPICE score gains in larger models do not justify the increased carbon footprint.

\noindent\textbf{Visual-QA: } On the Path-VQA dataset~\cite{he2020pathvqa}, \textbf{GPT-4o} achieves a BLEU score of \textbf{0.0037}(Table \ref{tab:visual_qa_scores}), matching the base Qwen-VL-2B model but significantly lower than the fine-tuned models. After fine-tuning, \textbf{Qwen-VL-7B} achieves a BLEU score of \textbf{0.0750}, with a \textbf{7,570\%} gain over its base model and a \textbf{1,927.03\%} gain compared to GPT-4o. The smaller model \textbf{Qwen-VL-2B} also performs well, achieving a BLEU score of \textbf{0.0592}, reflecting a \textbf{1,500\%} improvement over its base model and GPT-4o. In terms of emissions, \textbf{Qwen-VL-2B} emits \textbf{60.43g} compared to \textbf{212.76g} for \textbf{Llama-3.2-11B}, whose marginal BLEU improvements do not justify the increased carbon footprint. These results highlight the efficiency of smaller models like Qwen-VL-2B and Qwen-VL-7B in achieving substantial performance improvements with significantly lower environmental costs.

\begin{table}[H]
    \centering
    \begin{tabular}{l|c|c|c|c|c}
        \toprule
        \textbf{Model($M$)} & \boldmath{$L_r$} & \boldmath{$B_M$} \textbf{BLEU} & \boldmath{$F_T$} \textbf{BLEU} & \textbf{Gain (\boldmath{$F_T$} vs \boldmath{$B_M$})} & \textbf{Gain (\boldmath{$F_T$} vs \textbf{GPT-4o})} \\
        \midrule
        \multirow{3}{*}{Qwen-VL-2B} 
            & 8  & 0.0037 & 0.0552 & +1,391.89\% & +1,391.89\% \\
            & 16 & 0.0037 & 0.0592 & +1,500.00\% & +1,500.00\% \\
            & 32 & 0.0037 & 0.0587 & +1,486.49\% & +1,486.49\% \\
        \cmidrule(lr){1-6}
        \multirow{3}{*}{Qwen-VL-7B} 
            & 8  & 0.0010 & 0.0732 & +7,220.00\% & +5,538.46\% \\
            & 16 & 0.0010 & \textbf{0.0767} & \textbf{+7,570.00\%} & \textbf{+1972.03}\% \\
            & 32 & 0.0010 & 0.0750 & +7,400.00\% & +1927.02\% \\
        \cmidrule(lr){1-6}
        \multirow{3}{*}{Llama-3.2-11B} 
            & 8  & 0.0006 & 0.0014 & +133.33\% & \textcolor{red}{--62.16\%} \\
            & 16 & 0.0006 & 0.0017 & +183.33\% & \textcolor{red}{--54.05\%} \\
            & 32 & 0.0006 & 0.0017 & +183.33\% & \textcolor{red}{--54.05\%} \\
        \bottomrule
    \end{tabular}
    \vspace{2mm}
    \caption{Performance gain after fine-tuning compared to $B_M$ and GPT-4o for Visual QA task.}
    \label{tab:visual_qa_performance}
\end{table}

\subsubsection{Summarization and Test-to-SQL}

\textbf{Dialogue Summarization: } GPT-4o has a ROUGE-1($R_1$) score (Table \ref{tab:summarization_scores}) of \textbf{0.37} on SAMSum dataset\cite{gliwa-etal-2019-samsum}. Fine-tuned models outperform GPT-4o, with gains ranging from \textbf{13.51} \% to \textbf{37.84} \% as shown in the table \ref{tab:dialogue_summarization_performance}. Larger models like Qwen2.5-14B show higher gains(37.84 \%) over GPT-4o.

\begin{figure}[H]
    \centering
    \begin{minipage}[t]{0.48\linewidth}
        \centering
        \includegraphics[width=\linewidth]{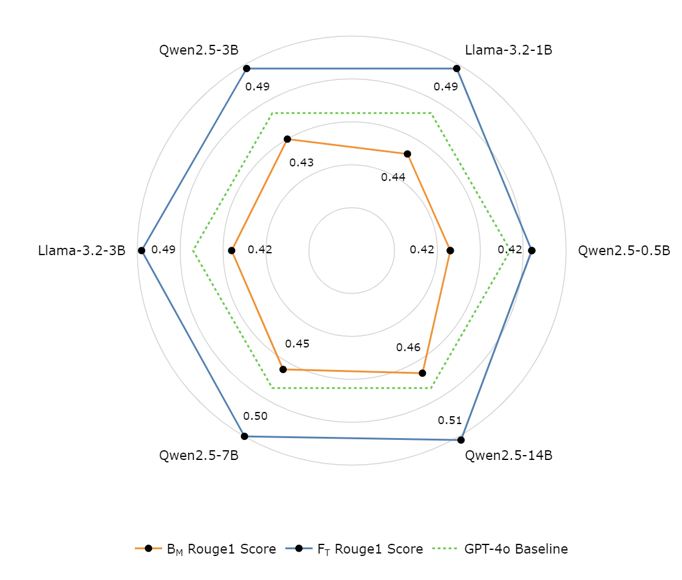}
        \caption{Comparison of ROUGE-1 scores for $B_M$, $F_T$, and GPT-4o in Dialogue Summarization. Fine-tuned models demonstrate substantial improvements over the base models and GPT-4o baseline, showcasing their efficacy in generating summaries with higher semantic relevance.}
        \label{fig:figure_summarization_performance}
    \end{minipage}
    \hfill
    \begin{minipage}[t]{0.48\linewidth}
        \centering
        \includegraphics[width=\linewidth]{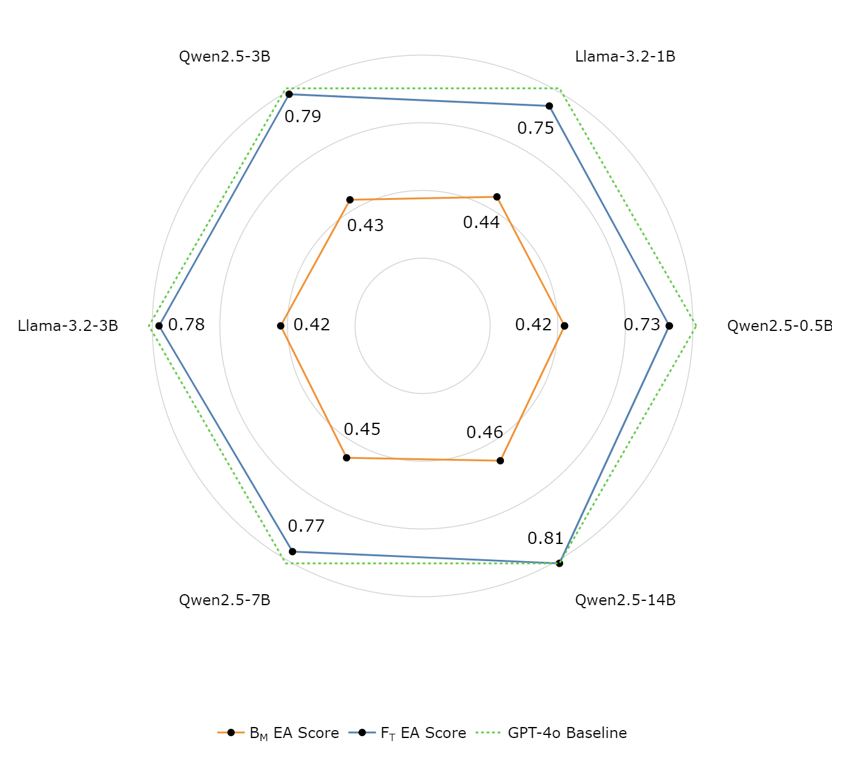}
        \caption{Comparison of Execution Accuracy (EA) scores for $B_M$, $F_T$, and GPT-4o in Text-to-SQL tasks. Fine-tuned models achieve significant accuracy gains over corresponding $B_M$.}
        \label{fig:figure_ttsql_performance}
    \end{minipage}
\end{figure}

All models exhibit significant improvements after fine-tuning compared to their base models, as shown in the chart \ref{fig:figure_summarization_performance}. The percentage gains range from \textbf{54.55 \%} (Qwen2.5-14B) to \textbf{+88.46 \%} (Llama-3.2-1B). Smaller models like Qwen2.5-0.5B and Llama-3.2-1B achieve substantial performance gains with lower carbon emissions of 25.30 gm and 37.68 gm, respectively. In table \ref{tab:dialogue_summarization_performance} and \ref{tab:text_to_sql_performance}, \textbf{Avg.} represents average metric score obtained by a model over $L_r$ and $Q_b$.

\begin{table}[H]
    \centering
    \begin{tabular}{l|c|c|c|c}
        \toprule
        \textbf{Model($M$)}  & \textbf{Avg. }(\boldmath{$B_M$}) \boldmath{$R_1$} & \textbf{Avg. } (\boldmath{$F_T$}) \boldmath{$R_1$} & \textbf{Gain (\boldmath{$F_T$} vs \boldmath{$B_M$})} & \textbf{Gain (\boldmath{$F_T$} vs \textbf{GPT-4o})} \\
        \midrule
        Qwen2.5-0.5B   & 0.23 & 0.42 & +82.61\% & +13.51\% \\
        Qwen2.5-3B      & 0.30 & 0.50 & +66.67\% & +35.14\% \\
        Qwen2.5-7B       & 0.32 & 0.50 & +56.25\% & +35.14\% \\
        Qwen2.5-14B    & \textbf{0.33} & \textbf{0.51} & +54.55\% & \textbf{+37.84\%} \\
        Llama-3.2-1B    & 0.26 & 0.49 & \textbf{+88.46\%} & +32.43\% \\
        Llama-3.2-3B   & 0.28 & 0.49 & +75.00\% & +32.43\% \\
        \bottomrule
    \end{tabular}
    \vspace{2mm}
    \caption{Performance gain after fine-tuning compared to $B_M$ and GPT-4o for the Dialogue Summarization task.}
    \label{tab:dialogue_summarization_performance}
\end{table}

\textbf{Text-To-SQL: }  GPT-4o has EA,VES = 0.80, 0.98 as mentioned in table \ref{tab:text_to_sql_scores} on synthetic\_text\_to\_sql dataset\cite{gretel-synthetic-text-to-sql-2024}. Qwen2.5-14B slightly surpasses GPT-4o's EA by \textbf{0.21 \%}. Other models have EA slightly below GPT-4o, with negative gains ranging from \textbf{-2.06 \%} to \textbf{-9.67 \%}. All models maintain high VES scores (Table \ref{tab:text_to_sql_performance}), closely matching or slightly exceeding GPT-4o's VES.

\begin{table}[H]
    \centering
    \begin{tabular}{l|c|c|c|c}
        \toprule
        \textbf{Model($M$)} & \textbf{Avg.} \boldmath{$(B_M, F_T)$} \textbf{EA} & \textbf{Avg.} \boldmath{$(B_M, F_T)$} \textbf{VES}  & \textbf{Gain (\boldmath{$F_T$} vs \boldmath{$B_M$}) EA} & \textbf{Gain (\boldmath{$F_T$} vs \textbf{GPT-4o}) EA} \\
        \midrule
        Qwen2.5-0.5B    & (0.42, 0.73) & (0.97, 0.98)  & +74.21\% & \textcolor{red}{--9.67\%} \\
        Qwen2.5-3B      & (0.43, 0.79) & (0.97, 0.98)  & +84.50\% & \textcolor{red}{--2.06\%} \\
        Qwen2.5-7B      & (0.45, 0.77) & (0.98, 0.98)  & +71.85\% & \textcolor{red}{--4.53\%} \\
        Qwen2.5-14B     & (0.46, 0.81) & (0.98, 0.98) & +76.45\% & \textcolor{Green}{\textbf{+0.21\%}} \\
        Llama-3.2-1B    & (0.45, 0.75) & (0.97, 0.98)  & +67.84\% & \textcolor{red}{--7.82\%} \\
        Llama-3.2-3B    & (0.41, 0.78) & (0.97, 0.98)  & \textbf{+88.36\%} & \textcolor{red}{--3.50\%} \\
        \bottomrule
    \end{tabular}
    \vspace{2mm}
    \caption{Performance gain after fine-tuning compared to $B_M$ and GPT-4o for the Text-to-SQL task. Negative gains are highlighted in red to indicate performance below GPT-4o.}
    \label{tab:text_to_sql_performance}
\end{table}

Qwen2.5-14B achieves the highest EA, slightly outperforming GPT-4o, but at the cost of more carbon emissions. Smaller models like Llama-3.2-1B and Qwen2.5-0.5B achieve substantial improvements over their base versions but still need to catch up to GPT-4o's EA.

Performance Consistency: There's less than 2\% variation across different quantization levels. Lower-bit quantization levels effectively maintain performance while reducing carbon emissions, promoting energy efficiency. Lower-bit quantization (4-bit or 8-bit) delivers nearly the same performance as higher-bit quantization (16-bit or 32-bit) while reducing carbon emissions by up to 10.6\%.

\subsection{Performance Vs Emission: The Trade-off}

This section examines the balance between model performance and carbon emissions across four key tasks: Dialogue Summarization, Image Captioning, Visual QA, and Text-to-SQL Conversion. By analyzing the mean data obtained from various configurations—including different LoRA ranks and quantization levels—we highlight how fine-tuning smaller models and leveraging lower-bit quantization can offer substantial performance gains with reduced environmental impact. Figure \ref{fig:greener_model} (appendix \ref{sec:experiments}) mentions emission efficiency of best performing fine-tuned model for each task.

% \begin{figure}[H]
%     \centering
%     \begin{minipage}[t]{0.68\linewidth}
%         \centering
%         \includegraphics[width=1.1\linewidth]{figures/results/ic_carbon.png}
%         \caption{Comparison of SPICE scores and carbon emissions for $B_M$, $F_T$ for Image Captioning.}
%         \label{fig:figure_image_caption_performance}
%     \end{minipage}
%     \hfill
%     \begin{minipage}[t]{0.68\linewidth}
%         \centering
%         \includegraphics[width=1.1
% \linewidth]{figures/results/vqa_carbon.png}
%         \caption{Comparison of BLEU scores and carbon emissions for $B_M$, $F_T$ for Visual QA.}
%         \label{fig:figure_visual_qa_performance}
%     \end{minipage}
% \end{figure}

\begin{figure}[t]
    \centering
    \includegraphics[width=0.8\linewidth]{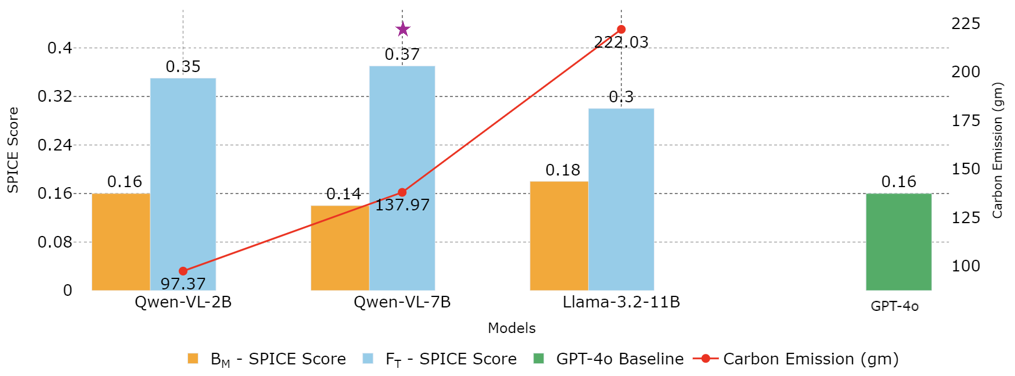}
    \caption{Comparison of SPICE scores and carbon emissions for $B_M$, $F_T$ for Image Captioning.}
    \label{fig:carbon_image_caption_performance}
\end{figure}

\begin{figure}[t]
    \centering
    \includegraphics[width=0.8\linewidth]{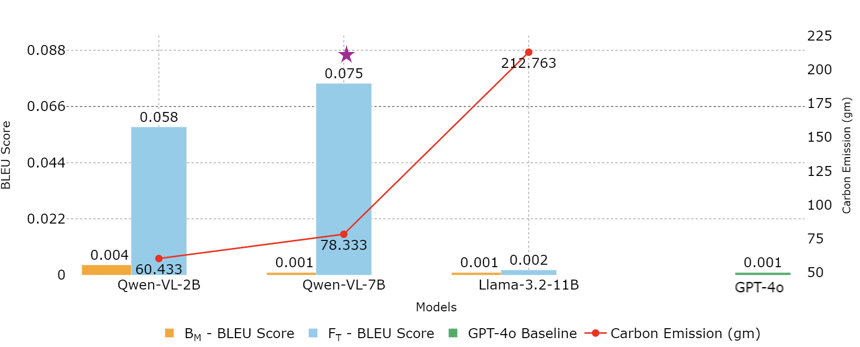}
    \caption{Comparison of BLEU scores and carbon emissions for $B_M$, $F_T$ for Visual QA.}
    \label{fig:carbon_visual_qa_performance}
\end{figure}

\textbf{Image Captioning:} A shown in the Plot \ref{fig:carbon_image_caption_performance} moving from Qwen-VL-2B to 3x sized Qwen-VL-7B results in a SPICE score increase of only 5.56\% (from 0.36 to 0.38), while carbon emissions increase by 41.30\% (from 97.8 gm to 138.2 gm). Further increasing to Llama-3.2-11B, the SPICE score decreases to 0.31, despite carbon emissions increasing\ref{tab:spice_scores} by 126.49\% compared to Qwen-VL-2B. Qwen-VL-2B enhances performance by over 118\% (SPICE score from 0.16 to 0.35), achieving 94.6\% of the larger model's performance (0.35 vs. 0.37) while emitting 56\% less carbon emissions (97.37gm vs. 222.03gm for Llama-3.2-11B).

\textbf{Visual QA: }Qwen-VL-7B achieves a BLEU score of 0.0750, which is 30.16\% higher than Qwen-VL-2B's BLEU score of 0.0577. However, this comes with a 29.59\% increase in carbon emissions(Plot \ref{fig:carbon_visual_qa_performance}). Surprisingly, Llama3.2-11B emits +252.03\% more carbon (Table \ref{tab:visual_qa_scores}) than Qwen-VL-2B while there is a huge decline in performance by 97\%. Fine-tuning Qwen-VL-2B boosts performance by over 1,460\% (BLEU score from 0.0037 to 0.0577), achieving 77\% of the larger model Qwen-VL-7B's accuracy (0.0577 vs. 0.0750) with 23\% less carbon emissions (60.43 gm vs. 78.33 gm).

\begin{figure}[t]
    \centering
    \includegraphics[width=0.8\linewidth]{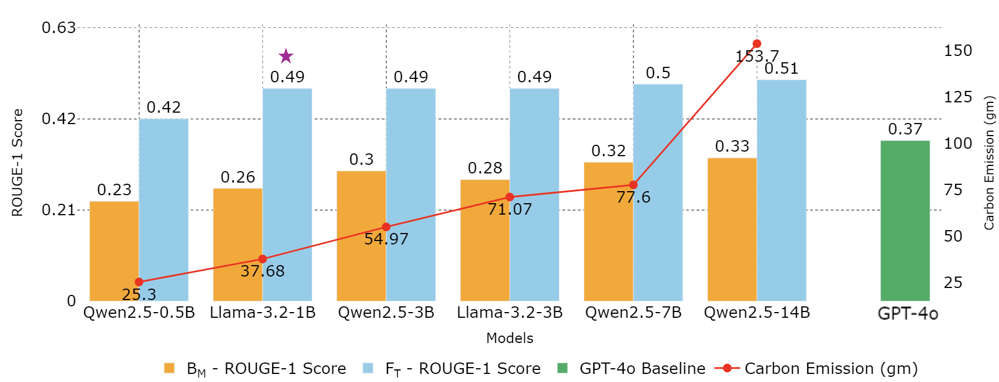}
    \caption{Comparison of ROUGE-1 scores and carbon emissions for $B_M$, $F_T$ for Dialogue Summarization.}
    \label{fig:figure_dialogue_summarization_performance}
\end{figure}

\begin{figure}[t]
    \centering
    \includegraphics[width=0.8\linewidth]{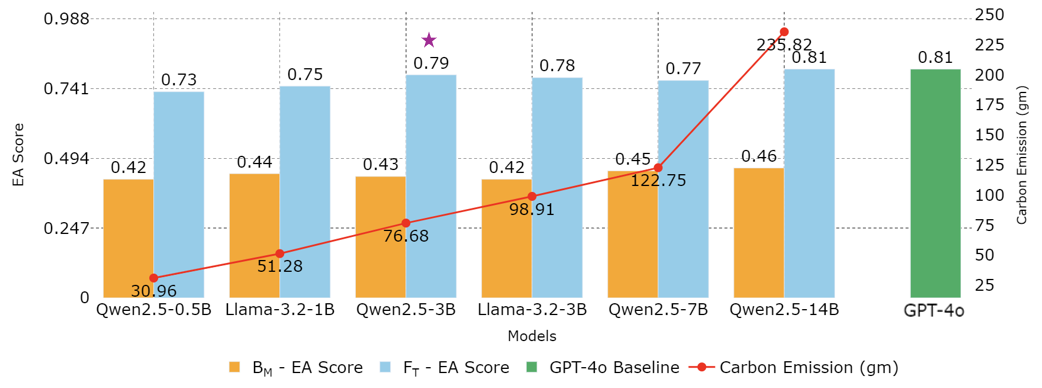}
    \caption{Comparison of EA scores and carbon emissions for $B_M$, $F_T$ for Text-to-SQL.}
    \label{fig:figure_text_to_sql_performance}
\end{figure}

\textbf{Dialogue Summarization: }Qwen2.5-0.5B emits 25.30 gm, while Qwen2.5-14B emits 153.70 gm, an increase of 507.51\% for only 4\% improvement in performance with 28x model size. Also, there is a marginal increase in ROUGE-1 score from 0.50 to 0.51(Plot  \ref{fig:figure_dialogue_summarization_performance}), only a 2\% improvement when moving from Qwen2.5-7B to Qwen2.5-14B results in a 98.09\% increase in carbon emissions( Table \ref{tab:summarization_scores}). Qwen2.5-0.5B improves performance by over 82\% (ROUGE-1 score from 0.23 to 0.42), achieving 98\% of the larger model's accuracy (compared to Qwen2.5-14B's 0.51) with 83\% less carbon emissions (25.30gm vs. 153.70gm). 

\textbf{Text-To-SQL: }Plot \ref{fig:figure_text_to_sql_performance} shows that smaller models like Qwen2.5-3B, Llama3.2-3B achieve EA (Table \ref{tab:text_to_sql_scores}) close to GPT-4o with lower emmisions compared to Qwen2.5-14B. The slight EA improvement of +0.21\% over GPT-4o by Qwen2.5-14B comes with a 661.72\% increase in emissions compared to Qwen2.5-0.5B.  Qwen2.5-0.5B enhances EA by over 74\% (from 0.42 to 0.7317), reaching 90\% of the largest model's performance (0.7317 vs. 0.8117) with 86.9\% less carbon emissions (30.96gm vs. 235.82gm).

The model marked with the `\textbf{{\textcolor{Purple}{$\star$}}}' symbol in plots \ref{fig:figure_image_caption_performance}, \ref{fig:figure_visual_qa_performance}, \ref{fig:figure_dialogue_summarization_performance}, and \ref{fig:figure_text_to_sql_performance} represents the optimal model, achieving maximum performance with minimum carbon emissions. This model can be considered as the \textit{sweet spot} of the emission versus performance trade-off. 

\subsection{Emission per unit gain}

To analyze the performance gain comprehensively, we consider various quantization levels \( Q_b \) and LoRA ranks \( L_r \) for a model $M$. The average percentage gain \( \bar{G}_{M,\mu} \) for metric \( \mu \) across all configurations is calculated as:

\begin{equation}
    \bar{G}_{M,\mu} = \frac{\sum\limits_{\forall Q_{b}} \sum\limits_{\forall L_{r}} \text{G}_{M,\mu}(F_T, B_M)}{|Q_{b}| \cdot |L_{r}|} 
    \label{eq:avg_gain}
\end{equation}

\noindent where:
\begin{itemize}
    \item \(\text{G}_{M,\mu}(F_T, B_M)\): The percentage gain for metric \( \mu \) between the fine-tuned model (\( F_T \)) and the base model (\( B_M \)) for all combination of \( Q_b \) and \( L_r \).
    \item \( Q_b = \{4\text{-bit}, 8\text{-bit}\} \): The set of quantization bit considered.
    \item \( L_r = \{4, 8, 16, 32\} \): The set of LoRA ranks considered.
    \item \( |Q_b| \) and \( |L_r| \): The cardinalities of sets \( Q_b \) and \( L_r \), respectively.
\end{itemize}

We define the \textbf{emission per unit percentage gain} \( G_{M,\mu}^{o} \) as the ratio of the total carbon emissions to the total percentage gain across all configurations as

\begin{align}
    G^{o}_{M,\mu} &= \frac{\sum\limits_{\forall Q_b} \sum\limits_{\forall L_r} C_E}{\bar{G}_{M,\mu} \cdot |Q_b| \cdot |L_r|} \label{eq:unit_gain} \\
    \vspace{3mm} \nonumber \\
                  &= \frac{\sum\limits_{\forall Q_b} \sum\limits_{\forall L_r} C_E}{|Q_b| \cdot |L_r|} 
                  \times \frac{|Q_b| \cdot |L_r|}{\sum\limits_{\forall Q_b} \sum\limits_{\forall L_r} G_{M,\mu}(F_T, B_M)} \label{eq:expanded_gain} \\
    \vspace{3mm} \nonumber \\
                  &= \frac{\sum\limits_{\forall Q_b} \sum\limits_{\forall L_r} C_E}{\sum\limits_{\forall Q_b} \sum\limits_{\forall L_r} G_{M, \mu}(F_T, B_M)} \label{eq:simplified_gain}
\end{align}

To normalize the unit percentage gain by the number of trainable parameters, we introduce \( T_p \), the number of \textbf{trainable parameters} (in millions) after applying LoRA. The \textbf{emission per unit percentage gain per million trainable parameters} \( G_{M,\mu,T_p}^{o} \), which is the \textbf{CEGI} (Carbon Efficient Gain Index) score, defined as

\begin{align}
   G^{o}_{M,\mu,T_p} &= \frac{G^{o}_{M,\mu} \times |L_r|}{\sum\limits_{\forall L_r} T_p} \label{eq:gain_per_param} \\
   \vspace{3mm} \nonumber \\
                     &= \frac{\bigg(\sum\limits_{\forall Q_b} \sum\limits_{\forall L_r} C_E\bigg) \times |L_r|}{\bigg(\sum\limits_{\forall Q_b} \sum\limits_{\forall L_r} G_{M,\mu}(F_T, B_M)\bigg) \times \sum\limits_{\forall L_r} T_p} \label{eq:gain_per_param_expanded}
\end{align}

\noindent where \( \sum\limits_{\forall L_r} T_p \): Total number of trainable parameters (in millions) summed over all LoRA ranks \( L_r \). Table \ref{tab:decomposed_params} in appendix section \ref{sec:experiments} provides trainable parameters for various model sizes and $L_r$\\

\noindent This index \textbf{CEGI} ( \( G_{M,\mu,T_p}^{o} \)) quantifies the carbon emission per unit percentage gain per million trainable parameters, providing a normalized measure to compare models of different sizes and configurations. The lower CEGI score, the better is the performance. By utilizing this metric, we can effectively compare the efficiency of different models and configurations regarding performance improvement and environmental impact.

\begin{table}[H]
    \centering
    \begin{tabular}{l|c|c|c|c}
        \toprule
        \textbf{Model($M$}) & \textbf{Train params} \boldmath{$\bigg(\sum_{\forall_{L_r} }T_p\bigg)$} & \boldmath{$\bar{C_E}$(gm)} & \textbf{\%Gain} \boldmath{$\bigg(\bar{G}_{M,\mu}\bigg)$} & \textbf{CEGI} \bigg( \( G_{M,\mu,T_p}^{o} \)\bigg)  \\
        \midrule
        Llama-3.2-11B & 3.92m & 222.03 & 66.67 & 1.18 \\
        Qwen-VL-2B    & 1.67m & 97.37  & 120.83 & 0.65 \\
        Qwen-VL-7B    & 3.12m & 137.97 & 166.67 & \textcolor{Green}{\textbf{0.36}} \\
        \bottomrule
    \end{tabular}
    \vspace{2mm}
    \caption{Comparison of \textbf{CEGI} scores across model variants for the image captioning task.}
    \label{tab:per_accuracy_image}
\end{table}

\textbf{Qwen-VL-7B} achieves the lowest \textbf{CEGI} value of \textbf{0.36}, indicating it is the most efficient model in terms of carbon emissions per unit percentage gain per million trainable parameters (Table \ref{tab:per_accuracy_image}). Despite having a moderate number of trainable parameters (\( 3.12 \) million), Qwen-VL-7B attains the highest percentage gain of \( 166.67\% \), demonstrating a solid balance between performance and efficiency. For the Image Captioning task, Qwen-VL-7B is the most efficient model, offering the best trade-off between performance gain and carbon emissions per million trainable parameters. Similar observations and conclusions can be made for Visual QA, Dialogue summarization, and Text-to-SQL tasks by referring to corresponding tables \ref{tab:per_accuracy_vision}, \ref{tab:per_accuracy_summ}, and \ref{tab:per_accuracy_sql} respectively.

\begin{table}[H]
    \centering
    \begin{tabular}{l|c|c|c|c}
        \toprule
       \textbf{Model($M$}) & \textbf{Train params} \boldmath{$\bigg(\sum_{\forall_{L_r} }T_p\bigg)$} & \boldmath{$\bar{C_E}$(gm)} & \textbf{\%Gain} \boldmath{$\bigg(\bar{G}_{M,\mu}\bigg)$} & \textbf{CEGI} \bigg( \( G_{M,\mu,T_p}^{o} \)\bigg) \\
        \midrule
        Llama-3.2-11B & 3.92m  & 212.76 & 166.67 & 0.489 \\
        Qwen-VL-2B    & 1.67m  & 60.43  & 1459.46 & 0.034 \\
        Qwen-VL-7B    & 3.12m & 78.33  & 7396.67 & \textcolor{Green}{\textbf{0.005}} \\
        \bottomrule
    \end{tabular}
    \vspace{2mm}
    \caption{Comparison of \textbf{CEGI} scores across model variants for the Visual-QA task.}
    \label{tab:per_accuracy_vision}
\end{table}

\begin{table}[H]
    \centering
    \begin{tabular}{l|c|c|c|c}
        \toprule
        \textbf{Model($M$}) & \textbf{Train params} \boldmath{$\bigg(\sum_{\forall_{L_r} }T_p\bigg)$} & \boldmath{$\bar{C_E}$(gm)} & \textbf{\%Gain} \boldmath{$\bigg(\bar{G}_{M,\mu}\bigg)$} & \textbf{CEGI} \bigg( \( G_{M,\mu,T_p}^{o} \)\bigg) \\
        \midrule
        Llama-3.2-1B    & 0.59m  & 37.68  & 87.82  & \textcolor{Green}{\textbf{0.99}} \\
        Llama-3.2-3B    & 1.02m  & 71.07  & 74.40   & 1.25 \\
        Qwen2.5-0.5B    & 0.36m  & 25.30   & 84.06  & 1.03 \\
        Qwen2.5-14B     & 2.21m  & 153.70  & 55.56  & 1.69 \\
        Qwen2.5-3B      & 1.02m  & 54.97  & 65.00     & 1.11 \\
        Qwen2.5-7B      & 1.56m  & 77.60   & 57.29  & 1.18 \\
        \bottomrule
    \end{tabular}
    \vspace{2mm}
    \caption{Comparison of \textbf{CEGI} scores across model variants for the dialogue summarization task.}
    \label{tab:per_accuracy_summ}
\end{table}

\begin{table}[H]
    \centering
    \begin{tabular}{l|c|c|c|c}
        \toprule
        \textbf{Model($M$}) & \textbf{Train params} \boldmath{$\bigg(\sum_{\forall_{L_r} }T_p\bigg)$} & \boldmath{$\bar{C_E}$(gm)} & \textbf{\%Gain} \boldmath{$\bigg(\bar{G}_{M,\mu}\bigg)$} & \textbf{CEGI} \bigg( \( G_{M,\mu,T_p}^{o} \)\bigg) \\
        \midrule
        Llama-3.2-1B    & 0.59m  & 51.28  & 67.80   & 1.77 \\
        Llama-3.2-3B    & 1.02m  & 98.91  & 88.39  & 1.53 \\
        Qwen2.5-0.5B    & 0.36m  & 30.96  & 74.21  & 1.43 \\
        Qwen2.5-14B     & 2.21m  & 235.82 & 76.45  & 1.89 \\
        Qwen2.5-3B      & 1.02m  & 76.68  & 84.50   & \textcolor{Green}{\textbf{1.22}} \\
        Qwen2.5-7B      & 1.56m  & 122.75 & 71.85  & 1.49 \\
        \bottomrule
    \end{tabular}
    \vspace{2mm}
    \caption{Comparison of \textbf{CEGI} scores across model variants for the Text-to-SQL task.}
    \label{tab:per_accuracy_sql}
\end{table}

Furthermore, the model with the lowest Carbon Efficient Gain Index (CEGI) score across all four tasks (highlighted in \textcolor{Green}{\textbf{green}} in tables \ref{tab:per_accuracy_image}, \ref{tab:per_accuracy_vision}, \ref{tab:per_accuracy_summ}, and\ref{tab:per_accuracy_sql}) consistently identifies the same model selected through human judgment, which is also marked with the `\textbf{{\textcolor{Purple}{$\star$}}}' symbol in plots  \ref{fig:carbon_image_caption_performance}, \ref{fig:carbon_visual_qa_performance}, \ref{fig:figure_dialogue_summarization_performance}, and \ref{fig:figure_text_to_sql_performance}. This alignment signifies that the CEGI index is \textbf{well-correlated with human judgment}, so it is a reliable and unified metric to evaluate the emission-performance trade-off. Thus, the  CEGI index is a robust mechanism for selecting the best-performing and eco-friendly model that balances high efficiency and sustainability. It simplifies decision-making by quantitatively capturing the trade-off between model performance and environmental impact, guiding the selection of optimal models for diverse tasks.

\section{Conclusion}
This paper tries to address the crucial balance between model performance and environmental sustainability across four essential tasks: Image Captioning, Visual Question Answering, Dialogue Summarization, and Text-to-SQL Conversion. To quantify this trade-off, we introduced a novel index, the \textbf{CEGI} (\textit{Carbon Efficient Emission Index}). We believe this index significantly contributes to the field, allowing us to normalize and compare different models and configurations effectively.
Our findings reveal a surprising insight: fine-tuning smaller models can lead to significant performance improvements, often rivaling or surpassing larger models. This approach boosts accuracy and substantially reduces carbon emissions, providing a sustainable alternative to resource-intensive larger models.

We also discovered that lower-bit quantization techniques can significantly reduce energy consumption without compromising and sometimes even improving model performance. This finding challenges the conventional wisdom that larger models are always superior.
Our analysis indicates that larger models' incremental gains in accuracy may not justify their disproportionately higher environmental impact, especially when considering the normalized efficiency metric. By fine-tuning smaller models and utilizing efficient quantization techniques, we can achieve a harmonious balance between performance and sustainability. This has profound implications for developing and deploying machine learning models, paving the way for a more environmentally responsible AI future.

\subsection*{Limitations and Future Directions}
While our study offers valuable insights, it is essential to acknowledge its limitations. Our experiments were conducted on specific datasets and tasks, which may not fully represent the diverse range of real-world applications. Additionally, our focus on Qwen and Llama variants may represent the broader landscape of model architectures.
Future research could explore a broader range of models and tasks to validate and extend our conclusions. Investigating the impact of other efficiency techniques, such as model pruning and knowledge distillation, could further optimize the trade-off between performance and sustainability. Refining the CEGI metric to incorporate additional factors, like inference efficiency and deployment considerations, would provide a more comprehensive evaluation framework.

\section*{Acknowledgments}
This research was conducted at Synechron Innovation Labs, Bangalore. We would like to extend our heartfelt gratitude to \textbf{Hareesha Pattaje}, Managing Director - Technology, Synechron, for funding and supporting this work, and to \textbf{Akhlaque Khan}, Sr. Director - Technology, Synechron, for inspiring and guiding this exploration of the intersection between sustainability and AI through his valuable ideas and insights. Their contributions were instrumental in shaping the direction and success of this research.

%Bibliography
\bibliographystyle{unsrt}  
\bibliography{references}  

\begin{thebibliography}{10}

\bibitem{strubell2019energypolicyconsiderationsdeep}
Emma Strubell, Ananya Ganesh, and Andrew McCallum.
\newblock Energy and policy considerations for deep learning in nlp, 2019.

\bibitem{patterson2021carbonemissionslargeneural}
David Patterson, Joseph Gonzalez, Quoc Le, Chen Liang, Lluis-Miquel Munguia, Daniel Rothchild, David So, Maud Texier, and Jeff Dean.
\newblock Carbon emissions and large neural network training, 2021.

\bibitem{hu2021loralowrankadaptationlarge}
Edward~J. Hu, Yelong Shen, Phillip Wallis, Zeyuan Allen-Zhu, Yuanzhi Li, Shean Wang, Lu~Wang, and Weizhu Chen.
\newblock Lora: Low-rank adaptation of large language models, 2021.

\bibitem{brown2020languagemodelsfewshotlearners}
Tom~B. Brown, Benjamin Mann, Nick Ryder, Melanie Subbiah, Jared Kaplan, Prafulla Dhariwal, Arvind Neelakantan, Pranav Shyam, Girish Sastry, Amanda Askell, Sandhini Agarwal, Ariel Herbert-Voss, Gretchen Krueger, Tom Henighan, Rewon Child, Aditya Ramesh, Daniel~M. Ziegler, Jeffrey Wu, Clemens Winter, Christopher Hesse, Mark Chen, Eric Sigler, Mateusz Litwin, Scott Gray, Benjamin Chess, Jack Clark, Christopher Berner, Sam McCandlish, Alec Radford, Ilya Sutskever, and Dario Amodei.
\newblock Language models are few-shot learners, 2020.

\bibitem{anil2023palm2technicalreport}
Rohan Anil, Andrew~M Dai, Orhan Firat, Melvin Johnson, Dmitry Lepikhin, Alexandre Passos, Siamak Shakeri, Emanuel Taropa, Paige Bailey, Zhifeng Chen, et~al.
\newblock Palm 2 technical report.
\newblock {\em arXiv preprint arXiv:2305.10403}, 2023.

\bibitem{9563954}
Neil~C. Thompson, Kristjan Greenewald, Keeheon Lee, and Gabriel~F. Manso.
\newblock Deep learning's diminishing returns: The cost of improvement is becoming unsustainable.
\newblock {\em IEEE Spectrum}, 58(10):50--55, 2021.

\bibitem{henderson2022systematicreportingenergycarbon}
Peter Henderson, Jieru Hu, Joshua Romoff, Emma Brunskill, Dan Jurafsky, and Joelle Pineau.
\newblock Towards the systematic reporting of the energy and carbon footprints of machine learning, 2022.

\bibitem{wu2022sustainableaienvironmentalimplications}
Carole-Jean Wu, Ramya Raghavendra, Udit Gupta, Bilge Acun, Newsha Ardalani, Kiwan Maeng, Gloria Chang, Fiona~Aga Behram, James Huang, Charles Bai, Michael Gschwind, Anurag Gupta, Myle Ott, Anastasia Melnikov, Salvatore Candido, David Brooks, Geeta Chauhan, Benjamin Lee, Hsien-Hsin~S. Lee, Bugra Akyildiz, Maximilian Balandat, Joe Spisak, Ravi Jain, Mike Rabbat, and Kim Hazelwood.
\newblock Sustainable ai: Environmental implications, challenges and opportunities, 2022.

\bibitem{dodge2022measuringcarbonintensityai}
Jesse Dodge, Taylor Prewitt, Remi Tachet~Des Combes, Erika Odmark, Roy Schwartz, Emma Strubell, Alexandra~Sasha Luccioni, Noah~A. Smith, Nicole DeCario, and Will Buchanan.
\newblock Measuring the carbon intensity of ai in cloud instances, 2022.

\bibitem{budennyy2022eco2aicarbonemissionstracking}
Semen Budennyy, Vladimir Lazarev, Nikita Zakharenko, Alexey Korovin, Olga Plosskaya, Denis Dimitrov, Vladimir Arkhipkin, Ivan Oseledets, Ivan Barsola, Ilya Egorov, Aleksandra Kosterina, and Leonid Zhukov.
\newblock Eco2ai: carbon emissions tracking of machine learning models as the first step towards sustainable ai, 2022.

\bibitem{eco2AI}
S.~A. Budennyy, V.~D. Lazarev, N.~N. Zakharenko, A.~N. Korovin, O.~A. Plosskaya, D.~V. Dimitrov, V.~S. Akhripkin, I.~V. Pavlov, I.~V. Oseledets, I.~S. Barsola, I.~V. Egorov, A.~A. Kosterina, and L.~E. Zhukov.
\newblock eco2ai: Carbon emissions tracking of machine learning models as the first step towards sustainable ai.
\newblock {\em Doklady Mathematics}, January 2023.

\bibitem{li2018measuringintrinsicdimensionobjective}
Chunyuan Li, Heerad Farkhoor, Rosanne Liu, and Jason Yosinski.
\newblock Measuring the intrinsic dimension of objective landscapes, 2018.

\bibitem{aghajanyan-etal-2021-intrinsic}
Armen Aghajanyan, Sonal Gupta, and Luke Zettlemoyer.
\newblock Intrinsic dimensionality explains the effectiveness of language model fine-tuning.
\newblock In Chengqing Zong, Fei Xia, Wenjie Li, and Roberto Navigli, editors, {\em Proceedings of the 59th Annual Meeting of the Association for Computational Linguistics and the 11th International Joint Conference on Natural Language Processing (Volume 1: Long Papers)}, pages 7319--7328, Online, August 2021. Association for Computational Linguistics.

\bibitem{mscoco}
Tsung-Yi Lin, Michael Maire, Serge Belongie, James Hays, Pietro Perona, Deva Ramanan, Piotr Doll{\'a}r, and C.~Lawrence Zitnick.
\newblock Microsoft coco: Common objects in context.
\newblock In David Fleet, Tomas Pajdla, Bernt Schiele, and Tinne Tuytelaars, editors, {\em Computer Vision -- ECCV 2014}, pages 740--755, Cham, 2014. Springer International Publishing.

\bibitem{9999450}
Bo~Liu, Li-Ming Zhan, Li~Xu, and Xiao-Ming Wu.
\newblock Medical visual question answering via conditional reasoning and contrastive learning.
\newblock {\em IEEE Transactions on Medical Imaging}, 42(5):1532--1545, 2023.

\bibitem{biten2019scenetextvisualquestion}
Ali~Furkan Biten, Ruben Tito, Andres Mafla, Lluis Gomez, Marçal Rusiñol, Ernest Valveny, C.~V. Jawahar, and Dimosthenis Karatzas.
\newblock Scene text visual question answering, 2019.

\bibitem{NIPS2017_3f5ee243}
Ashish Vaswani, Noam Shazeer, Niki Parmar, Jakob Uszkoreit, Llion Jones, Aidan~N Gomez, \L~ukasz Kaiser, and Illia Polosukhin.
\newblock Attention is all you need.
\newblock In I.~Guyon, U.~Von Luxburg, S.~Bengio, H.~Wallach, R.~Fergus, S.~Vishwanathan, and R.~Garnett, editors, {\em Advances in Neural Information Processing Systems}, volume~30. Curran Associates, Inc., 2017.

\bibitem{9157506}
Yingwei Pan, Ting Yao, Yehao Li, and Tao Mei.
\newblock X-linear attention networks for image captioning.
\newblock In {\em 2020 IEEE/CVF Conference on Computer Vision and Pattern Recognition (CVPR)}, pages 10968--10977, 2020.

\bibitem{Herdade2019ImageCT}
Sim{\~a}o Herdade, Armin Kappeler, Kofi Boakye, and Joao Soares.
\newblock Image captioning: Transforming objects into words.
\newblock In {\em Neural Information Processing Systems}, 2019.

\bibitem{guo-etal-2019-star}
Qipeng Guo, Xipeng Qiu, Pengfei Liu, Yunfan Shao, Xiangyang Xue, and Zheng Zhang.
\newblock Star-transformer.
\newblock In Jill Burstein, Christy Doran, and Thamar Solorio, editors, {\em Proceedings of the 2019 Conference of the North {A}merican Chapter of the Association for Computational Linguistics: Human Language Technologies, Volume 1 (Long and Short Papers)}, pages 1315--1325, Minneapolis, Minnesota, June 2019. Association for Computational Linguistics.

\bibitem{radford2021learningtransferablevisualmodels}
Alec Radford, Jong~Wook Kim, Chris Hallacy, Aditya Ramesh, Gabriel Goh, Sandhini Agarwal, Girish Sastry, Amanda Askell, Pamela Mishkin, Jack Clark, Gretchen Krueger, and Ilya Sutskever.
\newblock Learning transferable visual models from natural language supervision, 2021.

\bibitem{gliwa-etal-2019-samsum}
Bogdan Gliwa, Iwona Mochol, Maciej Biesek, and Aleksander Wawer.
\newblock {SAMS}um corpus: A human-annotated dialogue dataset for abstractive summarization.
\newblock In Lu~Wang, Jackie Chi~Kit Cheung, Giuseppe Carenini, and Fei Liu, editors, {\em Proceedings of the 2nd Workshop on New Frontiers in Summarization}, pages 70--79, Hong Kong, China, November 2019. Association for Computational Linguistics.

\bibitem{chen-etal-2021-dialogsum}
Yulong Chen, Yang Liu, Liang Chen, and Yue Zhang.
\newblock {D}ialog{S}um: {A} real-life scenario dialogue summarization dataset.
\newblock In Chengqing Zong, Fei Xia, Wenjie Li, and Roberto Navigli, editors, {\em Findings of the Association for Computational Linguistics: ACL-IJCNLP 2021}, pages 5062--5074, Online, August 2021. Association for Computational Linguistics.

\bibitem{zhu-etal-2021-mediasum}
Chenguang Zhu, Yang Liu, Jie Mei, and Michael Zeng.
\newblock {M}edia{S}um: A large-scale media interview dataset for dialogue summarization.
\newblock In Kristina Toutanova, Anna Rumshisky, Luke Zettlemoyer, Dilek Hakkani-Tur, Iz~Beltagy, Steven Bethard, Ryan Cotterell, Tanmoy Chakraborty, and Yichao Zhou, editors, {\em Proceedings of the 2021 Conference of the North American Chapter of the Association for Computational Linguistics: Human Language Technologies}, pages 5927--5934, Online, June 2021. Association for Computational Linguistics.

\bibitem{yu-etal-2018-spider}
Tao Yu, Rui Zhang, Kai Yang, Michihiro Yasunaga, Dongxu Wang, Zifan Li, James Ma, Irene Li, Qingning Yao, Shanelle Roman, Zilin Zhang, and Dragomir Radev.
\newblock {S}pider: A large-scale human-labeled dataset for complex and cross-domain semantic parsing and text-to-{SQL} task.
\newblock In Ellen Riloff, David Chiang, Julia Hockenmaier, and Jun{'}ichi Tsujii, editors, {\em Proceedings of the 2018 Conference on Empirical Methods in Natural Language Processing}, pages 3911--3921, Brussels, Belgium, October-November 2018. Association for Computational Linguistics.

\bibitem{lee-etal-2021-kaggledbqa}
Chia-Hsuan Lee, Oleksandr Polozov, and Matthew Richardson.
\newblock {K}aggle{DBQA}: Realistic evaluation of text-to-{SQL} parsers.
\newblock In Chengqing Zong, Fei Xia, Wenjie Li, and Roberto Navigli, editors, {\em Proceedings of the 59th Annual Meeting of the Association for Computational Linguistics and the 11th International Joint Conference on Natural Language Processing (Volume 1: Long Papers)}, pages 2261--2273, Online, August 2021. Association for Computational Linguistics.

\bibitem{liao2022artbench}
Peiyuan Liao, Xiuyu Li, Xihui Liu, and Kurt Keutzer.
\newblock The artbench dataset: Benchmarking generative models with artworks.
\newblock {\em arXiv preprint arXiv:2206.11404}, 2022.

\bibitem{he2020pathvqa}
Xuehai He, Yichen Zhang, Luntian Mou, Eric Xing, and Pengtao Xie.
\newblock Pathvqa: 30000+ questions for medical visual question answering.
\newblock {\em arXiv preprint arXiv:2003.10286}, 2020.

\bibitem{papineni2002bleu}
Kishore Papineni, Salim Roukos, Todd Ward, and Wei-Jing Zhu.
\newblock Bleu: a method for automatic evaluation of machine translation.
\newblock In {\em Proceedings of the 40th annual meeting of the Association for Computational Linguistics}, pages 311--318, 2002.

\bibitem{lin-2004-rouge}
Chin-Yew Lin.
\newblock {ROUGE}: A package for automatic evaluation of summaries.
\newblock In {\em Text Summarization Branches Out}, pages 74--81, Barcelona, Spain, July 2004. Association for Computational Linguistics.

\bibitem{gretel-synthetic-text-to-sql-2024}
Yev Meyer, Marjan Emadi, Dhruv Nathawani, Lipika Ramaswamy, Kendrick Boyd, Maarten Van~Segbroeck, Matthew Grossman, Piotr Mlocek, and Drew Newberry.
\newblock {Synthetic-Text-To-SQL}: A synthetic dataset for training language models to generate sql queries from natural language prompts, April 2024.

\bibitem{DBLP:journals/corr/AndersonFJG16}
Peter Anderson, Basura Fernando, Mark Johnson, and Stephen Gould.
\newblock {SPICE:} semantic propositional image caption evaluation.
\newblock {\em CoRR}, abs/1607.08822, 2016.

\end{thebibliography}

\newpage
\appendix

\section{Dataset}
\label{sec:dataset}

\subsection{artbench-pd-256x256}
This dataset is a subset of public domain images from ArtBench, focusing on historical art styles like Art Nouveau and Ukiyo-e. The images are captioned using Florence-2-large.

\begin{table}[H]
    \centering
    \resizebox{\textwidth}{!}{%
    \begin{tabular}{|m{0.15\textwidth}|m{0.85\textwidth}|}
        \hline
        \textbf{Image} & \textbf{Caption} \\ 
        \hline
        \includegraphics[width=0.1\textwidth]{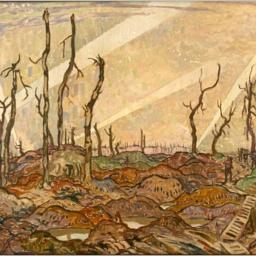} & 
        The image shows a painting of a landscape with dead trees in the foreground, painted by Vincent Van Gogh in 1889. The painting is titled "The Landscape with Dead Trees" and was created in 1888. \\ 
        \hline
        \includegraphics[width=0.1\textwidth]{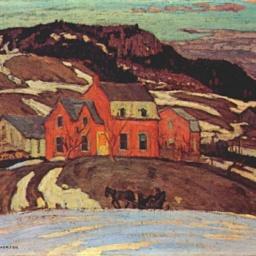} & 
        The image shows a painting of a red house in the middle of a snowy field, with a horse cart in front of it. The house is surrounded by trees and hills, and the sky is visible at the top of the painting. \\ 
        \hline
    \end{tabular}}
    \caption{Samples from Artbench dataset.}
    \label{tab:artbench_table}
\end{table}

\subsection{PathVQA}
PathVQA is a dataset designed for training and evaluating Medical Visual Question Answering (VQA) systems using pathology images. It features both open-ended and binary "yes/no" questions. The dataset is sourced from two publicly available pathology textbooks, Textbook of Pathology and Basic Pathology, as well as the Pathology Education Information Resource (PEIR) from the digital library.

\begin{table}[H]
    \centering
    \resizebox{\textwidth}{!}{%
    \begin{tabular}{|m{0.15\textwidth}|m{0.55\textwidth}|m{0.30\textwidth}|}
        \hline
        \textbf{Image} & \textbf{Question} & \textbf{Answer} \\ 
        \hline
        \includegraphics[width=0.1\textwidth]{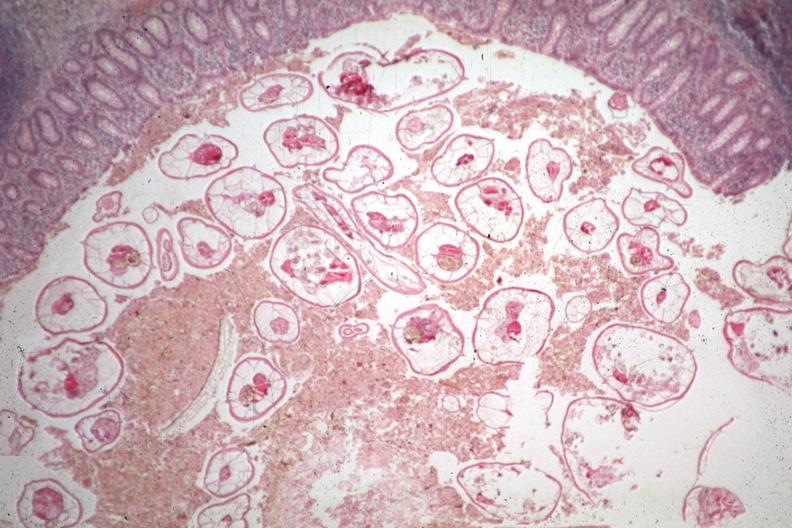} & 
        What does this image show? & Typical excellent pinworm \\ 
        \hline
        \includegraphics[width=0.1\textwidth]{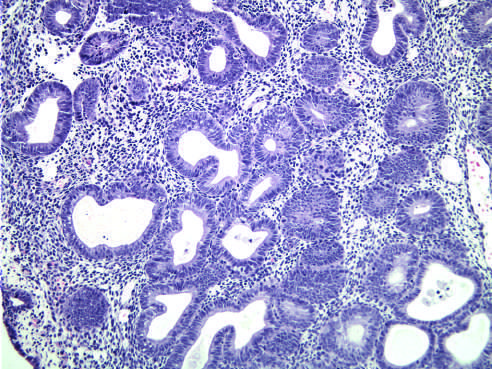} & 
        How is hyperplasia without atypia characterized? & By nests of closely packed glands \\ 
        \hline
    \end{tabular}}
    \caption{Samples from PathVQA dataset.}
    \label{tab:pathvqa_table}
\end{table}

\subsection{SAMSum}

The SAMSum dataset comprises messenger-style conversations written by English-speaking linguists to reflect real-life topics and styles, ranging from informal to formal as shown in Table~\ref{tab:samsum_table}. These conversations, featuring slang, emoticons, and typos, are annotated with concise third-person summaries that capture the key points discussed.

\begin{table}[H]
    \centering
    \resizebox{\textwidth}{!}{%
    \begin{tabular}{|m{0.65\textwidth}|m{0.35\textwidth}|}
        \hline
        \textbf{Dialogue} & \textbf{Summary} \\ 
        \hline
        Tim: Hi, what's up? Kim: Bad mood tbh, I was going to do lots of stuff but ended up procrastinating Tim: What did you plan on doing? Kim: Oh you know, uni stuff and unfucking my room Kim: Maybe tomorrow I'll move my ass and do everything Kim: We were going to defrost a fridge so instead of shopping I'll eat some defrosted veggies Tim: For doing stuff I recommend Pomodoro technique where u use breaks for doing chores Tim: It really helps Kim: thanks, maybe I'll do that Tim: I also like using post-its in kaban style & Kim may try the Pomodoro technique recommended by Tim to get more stuff done. \\ 
        \hline
        Sam: hey overheard rick say something Sam: i don't know what to do :-/ Naomi: what did he say?? Sam: he was talking on the phone with someone Sam: i don't know who Sam: and he was telling them that he wasn't very happy here Naomi: damn!!! Sam: he was saying he doesn't like being my roommate Naomi: wow, how do you feel about it? Sam: i thought i was a good roommate Sam: and that we have a nice place Naomi: that's true man!!! Naomi: i used to love living with you before i moved in with my boyfriend Naomi: i don't know why he's saying that Sam: what should i do??? Naomi: honestly if it's bothering you that much you should talk to him Naomi: see what's going on Sam: i don't want to get in any kind of confrontation though Sam: maybe I'll just let it go Sam: and see how it goes in the future Naomi: it's your choice Sam Naomi: if I were you I would just talk to him and clear the air & 
        Sam is confused because he overheard Rick complaining about him as a roommate. Naomi thinks Sam should talk to Rick. Sam is not sure what to do. \\
        \hline
    \end{tabular}}
    \caption{Samples from SAMSum dataset.}
    \label{tab:samsum_table}
\end{table}

\vspace{1cm} % Space between sections

\subsection{synthetic\_text\_to\_sql}

This dataset offers high-quality synthetic Text-to-SQL samples (Table~\ref{tab:textsql_table}), covering 100 domains with varying SQL complexities and natural language explanations. It supports a wide range of SQL tasks and is one of the largest and most diverse synthetic Text-to-SQL datasets, designed for advanced model training and research.

\begin{table}[H]
    \centering
    \resizebox{\textwidth}{!}{%
    \begin{tabular}{|m{0.20\textwidth}|m{0.40\textwidth}|m{0.40\textwidth}|}
        \hline
        \textbf{SQL Prompt} & \textbf{Context} & \textbf{SQL} \\ 
        \hline
        What is the total volume of timber sold by each salesperson, sorted by salesperson? & 
        \raggedright CREATE TABLE salesperson (salesperson\_id INT, name TEXT, region TEXT); 
        INSERT INTO salesperson (salesperson\_id, name, region) VALUES (1, 'John Doe', 'North'), (2, 'Jane Smith', 'South'); 
        CREATE TABLE timber\_sales (sales\_id INT, salesperson\_id INT, volume REAL, sale\_date DATE); 
        INSERT INTO timber\_sales (sales\_id, salesperson\_id, volume, sale\_date) VALUES 
        (1, 1, 120, '2021-01-01'), (2, 1, 150, '2021-02-01'), (3, 2, 180, '2021-01-01') & 
        SELECT salesperson\_id, name, SUM(volume) AS total\_volume FROM timber\_sales 
        JOIN salesperson ON timber\_sales.salesperson\_id = salesperson.salesperson\_id 
        GROUP BY salesperson\_id, name ORDER BY total\_volume DESC \\ 
        \hline
        List all the unique equipment types and their corresponding total maintenance frequency from the equipment\_maintenance table. & 
        CREATE TABLE equipment\_maintenance (equipment\_type VARCHAR(255), maintenance\_frequency INT); 
        INSERT INTO equipment\_maintenance (equipment\_type, maintenance\_frequency) VALUES 
        ('Type A', 10), ('Type B', 5), ('Type C', 12); & 
        SELECT equipment\_type, SUM(maintenance\_frequency) AS total\_maintenance\_frequency 
        FROM equipment\_maintenance GROUP BY equipment\_type \\ 
        \hline
    \end{tabular}}
    \caption{Samples from Text-to-SQL dataset.}
    \label{tab:textsql_table}
\end{table}

\newpage
\section{Prompts}
\label{sec:prompts}
This section contains task-specific prompts designed for performing inference and fine-tuning across the base and fine-tuned versions of Qwen-2.5, Qwen-VL, LLaMa-3.2. The prompts are tailored to handle specific tasks and optimize performance during , and GPT-4o zero-shot inference, ensuring high adaptability and relevance to diverse applications. For fine-tuning, the prompts are combined with ground-truth outputs (e.g., task\_prompt + ground\_truth) to enhance model performance. This dual-purpose design makes them versatile for both evaluation and training phases.

Temperature Set to \textbf{0} for fully deterministic outputs, ensuring consistent and predictable results without any randomness. Maximum Token Length: Capped at 1024 to ensure inference efficiency without truncation of outputs.

\begin{lstlisting}
# Image Caption task prompt

system_prompt = "You are an expert at creating clear and descriptive captions for historical artwork, focusing on main subjects and notable details."

prompt = """Generate a descriptive caption for the provided historical art image. 
Describe the main subjects, setting, and notable details visible in the image. Keep it concise and focused.

Return only the caption text."""

\end{lstlisting}

\begin{lstlisting}
# Visual QA prompt

system_prompt = "You are an expert at answering questions on pathology images, focusing on key details relevant to the question."

prompt = """Generate a concise answer to the provided ##question## about the pathology image.
Focus on key details and information directly visible in the image to answer accurately.

##question##: {question}
Return only the answer text."""
\end{lstlisting}

\begin{lstlisting}
# Dialogue Summarization

system_prompt = "You are an expert summarizer. Analyze the dialogue and provide a clear, concise summary capturing the main points and context."

prompt = """Analyze the conversation below and provide a concise, accurate summary that captures the key points and context.

### Dialogue:
{}
"""
\end{lstlisting}

\begin{lstlisting}
# Text-To-SQL prompt

system_prompt = "You are an advanced SQL query generator. Analyze the user's input and context to generate an efficient, accurate SQL query following best practices."

prompt = """Below is an input question asked by the user. Context is provided to clarify the user's question. Generate an SQL response based on the user's question.

### Input Question:
{}

### Context:
{}
"""
\end{lstlisting}

\newpage

\section{Experiments and results}
\label{sec:experiments}

Below table\ref{tab:text_sql} shows complete experiments results evaluating large language models (LLMs) and Small Language models on the text-to-sql data\ref{tab:textsql_table} set includes 14,000 training samples and 1,400 test samples,  their performance and computational efficiency in the text-to-SQL task. The models include Qwen2.5 (ranging from 0.5B to 14B parameters) and Llama-3.2 (1B and 3B parameters), fine-tuned across different configurations of quantization (4-bit and 8-bit) and ranks (4, 8, and 16). The primary metrics evaluated are EA and VES scores, which measure the models' ability to generate syntactically correct and semantically equivalent SQL queries. Furthermore, the carbon emissions (g) generated during the fine-tuning process and the comparison with GPT-4o's performance (serving as a baseline) are included to assess the environmental and computational cost.

\begin{table}[H]
    \centering
    \begin{tabular}{l|c|c|c|c|c|c|c}
        \toprule
        \textbf{Model($M$)} & \boldmath{$L_r$} & \boldmath{$Q_b$} & \boldmath{$C_E$}\textbf{(gm)} & \boldmath{$(B_{M},F_{T})$}\boldmath{$R_{1}$} & \boldmath{$(B_{M},F_{T})$}\boldmath{$R_{2}$} & \boldmath{($B_{M},F_{T})$}\boldmath{$R_{L}$} & \textbf{(GPT-4o)} \boldmath{$R_{(1, 2 , L)}$} \\
        \midrule
        % Qwen2.5-0.5B
        \multirow{6}{*}{Qwen2.5-0.5B} 
            & 4  & \multirow{3}{*}{4bit} & 25.2 & 0.23, 0.42 & 0.077, 0.16 & 0.17, 0.33 & \\
            & 8  &                       & 25.3 & 0.23, 0.41 & 0.077, 0.16 & 0.17, 0.33 & \\
            & 16 &                       & 25.4 & 0.23, 0.43 & 0.077, 0.17 & 0.17, 0.34 &\\
            \cmidrule(lr){2-7}
            & 4  & \multirow{3}{*}{8bit} & 25.2 & 0.23, 0.42 & 0.077, 0.16 & 0.17, 0.33 &\\
            & 8  &                       & 25.3 & 0.23, 0.43 & 0.077, 0.16 & 0.17, 0.33 &\\
            & 16 &                       & 25.4 & 0.23, 0.43 & 0.077, 0.17 & 0.17, 0.33 &\\
        \cmidrule(lr){1-7}
        % Qwen2.5-3B
        \multirow{6}{*}{Qwen2.5-3B} 
            & 4  & \multirow{3}{*}{4bit} & 56.4 & 0.3, 0.5 & 0.1, 0.24 & 0.23, 0.4 & \\
            & 8  &                       & 54.3 & 0.3, 0.5 & 0.1, 0.24 & 0.23, 0.4 &\\
            & 16 &                       & 55.3 & 0.3, 0.49 & 0.1, 0.24 & 0.23, 0.4 &\\
            \cmidrule(lr){2-7}
            & 4  & \multirow{3}{*}{8bit} & 54.3 & 0.3, 0.5 & 0.1, 0.24 & 0.23, 0.4 &\\
            & 8  &                       & 55.2 & 0.3, 0.49 & 0.1, 0.23 & 0.23, 0.4 &\\
            & 16 &                       & 54.3 & 0.3, 0.49 & 0.1, 0.24 & 0.23, 0.4 &\\
        \cmidrule(lr){1-7}
        % Qwen2.5-7B
        \multirow{6}{*}{Qwen2.5-7B} 
            & 4  & \multirow{3}{*}{4bit} & 78.2 & 0.32, 0.51 & 0.11, 0.24 & 0.24, 0.41 &\\
            & 8  &                       & 77.3 & 0.32, 0.5 & 0.11, 0.24 & 0.24, 0.4 &\\
            & 16 &                       & 78.3 & 0.32, 0.51 & 0.11, 0.25 & 0.24, 0.41 & \\
            \cmidrule(lr){2-7}
            & 4  & \multirow{3}{*}{8bit} & 77.2 & 0.32, 0.5 & 0.11, 0.24 & 0.24, 0.4 &\\
            & 8  &                       & 77.3 & 0.32, 0.5 & 0.11, 0.24 & 0.24, 0.41 & \multirow{6}{*}{0.37, 0.14, 0.28} \\
            & 16 &                       & 77.3 & 0.32, 0.5 & 0.11, 0.24 & 0.24, 0.41 &\\
        \cmidrule(lr){1-7}
        % Qwen2.5-14B
        \multirow{6}{*}{Qwen2.5-14B} 
            & 4  & \multirow{3}{*}{4bit} & 147.6 & 0.33, 0.51 & 0.12, 0.26 & 0.25, 0.42 &\\
            & 8  &                       & 147.8 & 0.33, 0.52 & 0.12, 0.26 & 0.25, 0.43 &\\
            & 16 &                       & 161.3 & 0.33, 0.51 & 0.12, 0.25 & 0.25, 0.42 &\\
            \cmidrule(lr){2-7}
            & 4  & \multirow{3}{*}{8bit} & 154.6 & 0.33, 0.51 & 0.12, 0.26 & 0.25, 0.43 &\\
            & 8  &                       & 152.5 & 0.33, 0.52 & 0.12, 0.26 & 0.25, 0.42 &\\
            & 16 &                       & 158.4 & 0.33, 0.51 & 0.12, 0.26 & 0.25, 0.43 &\\
        \cmidrule(lr){1-7}
        % Llama-3.2-1B
        \multirow{6}{*}{Llama-3.2-1B} 
            & 4  & \multirow{3}{*}{4bit} & 37.3 & 0.26, 0.49 & 0.08, 0.24 & 0.18, 0.4 &\\
            & 8  &                       & 38.1 & 0.26, 0.49 & 0.08, 0.23 & 0.18, 0.39 &\\
            & 16 &                       & 37.5 & 0.26, 0.48 & 0.08, 0.23 & 0.18, 0.4 &\\
            \cmidrule(lr){2-7}
            & 4  & \multirow{3}{*}{8bit} & 37.6 & 0.26, 0.49 & 0.08, 0.23 & 0.18, 0.4 &\\
            & 8  &                       & 39.2 & 0.26, 0.49 & 0.08, 0.23 & 0.18, 0.4 &\\
            & 16 &                       & 36.4 & 0.26, 0.49 & 0.08, 0.23 & 0.18, 0.4 &\\
        \cmidrule(lr){1-7}
        % Llama-3.2-3B
        \multirow{6}{*}{Llama-3.2-3B} 
            & 4  & \multirow{3}{*}{4bit} & 73.2 & 0.28, 0.5 & 0.1, 0.25 & 0.23, 0.41 &\\
            & 8  &                       & 71.8 & 0.28, 0.49 & 0.1, 0.24 & 0.23, 0.41 &\\
            & 16 &                       & 72.3 & 0.28, 0.49 & 0.1, 0.24 & 0.23, 0.4 &\\
            \cmidrule(lr){2-7}
            & 4  & \multirow{3}{*}{8bit} & 70.3 & 0.28, 0.5 & 0.1, 0.25 & 0.23, 0.41 &\\
            & 8  &                       & 70.2 & 0.28, 0.47 & 0.1, 0.23 & 0.23, 0.39 &\\
            & 16 &                       & 68.6 & 0.28, 0.48 & 0.1, 0.23 & 0.23, 0.39 &\\
        \bottomrule
    \end{tabular}
    \vspace{2mm}
    \caption{Comparison of model variants, quantization levels, carbon emissions, and ROUGE scores (Base, Fine-tuned) for various configurations.}
    
    \label{tab:rouge_scores}
\end{table}

\begin{table}[H]
    \centering
    \begin{tabular}{l|c|c|c|c|c|c}
        \toprule
        \textbf{Model($M$)} & \boldmath{$L_r$} &  \boldmath{$Q_b$} & \boldmath{$C_E$}\textbf{(gm)} & \boldmath{$(B_{M},F_{T})$}\textbf{EA} & \boldmath{$(B_{M},F_{T})$}\textbf{VES}  & \textbf{GPT-4o (EA, VES)} \\
        \midrule
        % Qwen2.5-0.5B
        \multirow{6}{*}{Qwen2.5-0.5B} 
            & 4  & \multirow{3}{*}{4bit} & 30.9  & 0.42, 0.73 & 0.97, 0.98 &  \\
            & 8  &                       & 30.96 & 0.42, 0.74 & 0.97, 0.98 &                         \\
             & 16 &                       & 30.98 & 0.42, 0.75 & 0.97, 0.98 &                         \\
            \cmidrule(lr){2-6}
             & 4  & \multirow{3}{*}{8bit} & 30.71 & 0.42, 0.72 & 0.97, 0.98 &                         \\
            & 8  &                       & 31.16 & 0.42, 0.73 & 0.97, 0.98 &                         \\
            & 16 &                       & 31.06 & 0.42, 0.72 & 0.97, 0.98 &                         \\
        \cmidrule(lr){1-6}
        % Qwen2.5-3B
        \multirow{6}{*}{Qwen2.5-3B} 
            & 4  & \multirow{3}{*}{4bit} & 78.11 & 0.43, 0.79 & 0.97, 0.98 & \\
            & 8  &                       & 77.94 & 0.43, 0.81 & 0.97, 0.99 & \\
            & 16 &                       & 75.36 & 0.43, 0.79 & 0.97, 0.98 & \\
            \cmidrule(lr){2-6}
            & 4  & \multirow{3}{*}{8bit} & 77.69 & 0.43, 0.79 & 0.97, 0.99 & \\
            & 8  &                       & 77.23 & 0.43, 0.79 & 0.97, 0.98 & \\
            & 16 &                       & 73.75 & 0.43, 0.79 & 0.97, 0.98 & \\
        \cmidrule(lr){1-6}
        % Qwen2.5-7B
        \multirow{6}{*}{Qwen2.5-7B} 
            & 4  & \multirow{3}{*}{4bit} & 123.2  & 0.45, 0.76 & 0.98, 0.99 & \\
            & 8  &                       & 122.67 & 0.45, 0.79 & 0.98, 0.98 & \\
            & 16 &                       & 122.84 & 0.45, 0.77 & 0.98, 0.99 & \\
            \cmidrule(lr){2-6}
            & 4  & \multirow{3}{*}{8bit} & 122.86 & 0.45, 0.78 & 0.98, 0.99 & \\
            & 8  &                       & 122.85 & 0.45, 0.77 & 0.98, 0.98 & \multirow{6}{*}{0.8, 0.98} \\ 
            & 16 &                       & 122.09 & 0.45, 0.77 & 0.98, 0.98 & \\
       \cmidrule(lr){1-6}
        % Qwen2.5-14B
        \multirow{6}{*}{Qwen2.5-14B} 
            & 4  & \multirow{3}{*}{4bit} & 237.31 & 0.46, 0.81 & 0.98, 0.99 & \\
            & 8  &                       & 234.2  & 0.46, 0.81 & 0.98, 0.98 & \\
            & 16 &                       & 235.9  & 0.46, 0.81 & 0.98, 0.99 & \\
            \cmidrule(lr){2-6}
            & 4  & \multirow{3}{*}{8bit} & 236.69 & 0.46, 0.82 & 0.98, 0.99 & \\
            & 8  &                       & 235.97 & 0.46, 0.81 & 0.98, 0.99 & \\
            & 16 &                       & 234.85 & 0.46, 0.81 & 0.98, 0.99 & \\
        \cmidrule(lr){1-6}
        % Llama-3.2-1B
        \multirow{6}{*}{Llama-3.2-1B} 
            & 4  & \multirow{3}{*}{4bit} & 53.35 & 0.44, 0.74 & 0.97, 0.98 & \\
            & 8  &                       & 48.56 & 0.44, 0.75 & 0.97, 0.98 & \\
            & 16 &                       & 49.72 & 0.44, 0.74 & 0.97, 0.98 & \\
            \cmidrule(lr){2-6}
            & 4  & \multirow{3}{*}{8bit} & 52.69 & 0.45, 0.75 & 0.97, 0.98 & \\
            & 8  &                       & 51.49 & 0.45, 0.75 & 0.97, 0.98 & \\
            & 16 &                       & 51.86 & 0.45, 0.75 & 0.97, 0.98 & \\
        \cmidrule(lr){1-6}
        % Llama-3.2-3B
        \multirow{6}{*}{Llama-3.2-3B} 
            & 4  & \multirow{3}{*}{4bit} & 107.03 & 0.41, 0.78 & 0.97, 0.99 & \\
            & 8  &                       & 96.25  & 0.41, 0.79 & 0.97, 0.98 & \\
            & 16 &                       & 96.94  & 0.41, 0.78 & 0.97, 0.98 & \\
            \cmidrule(lr){2-6}
            & 4  & \multirow{3}{*}{8bit} & 104.36 & 0.42, 0.77 & 0.98, 0.99 & \\
            & 8  &                       & 94.42  & 0.42, 0.78 & 0.98, 0.98 & \\
            & 16 &                       & 94.48  & 0.42, 0.79 & 0.98, 0.98 & \\
        \bottomrule
    \end{tabular}
    \vspace{2mm}
    \caption{Condensed comparison of model variants, quantization, EA, VES scores, and carbon emissions for text-to-SQL tasks.}
    \label{tab:text_sql}
\end{table}

\newpage
This table\ref{tab:decomposed_params} provides a detailed breakdown of decomposed parameters across various model sizes and LoRa matrix ranks, highlighting the number of trainable parameters in millions (M) for different configurations. These decomposed parameters result from applying technique low-rank adaptation (LoRA), where a large model's matrix operations are approximated using lower-rank matrices to reduce computational complexity while retaining.

\begin{table}[H]
    \centering
    \begin{tabular}{l|c|c}
        \toprule
        \textbf{Model Size} & \textbf{Lora rank}, \boldmath{$L_r$} & \textbf{Trainable Parameters}, \boldmath{$T_p$}\\
        \midrule
        \multirow{4}{*}{0.5B} & 4  & 0.18m \\
                              & 8  & 0.36m \\
                              & 16 & 0.72m \\
                              & 32 & 1.43m \\
        \midrule
        \multirow{4}{*}{1.0B} & 4  & 0.25m \\
                              & 8  & 0.51m \\
                              & 16 & 1.01m \\
                              & 32 & 2.02m \\
        \midrule
        \multirow{4}{*}{2.0B} & 4  & 0.36m \\
                              & 8  & 0.72m \\
                              & 16 & 1.43m \\
                              & 32 & 2.86m \\
        \midrule
        \multirow{4}{*}{3.0B} & 4  & 0.44m \\
                              & 8  & 0.88m \\
                              & 16 & 1.75m \\
                              & 32 & 3.51m \\
        \midrule
        \multirow{4}{*}{7.0B} & 4  & 0.67m \\
                              & 8  & 1.34m \\
                              & 16 & 2.68m \\
                              & 32 & 5.35m \\
        \midrule
        \multirow{4}{*}{11.0B} & 4  & 0.84m \\
                               & 8  & 1.68m \\
                               & 16 & 3.36m \\
                               & 32 & 6.71m \\
        \midrule
        \multirow{4}{*}{14.0B} & 4  & 0.95m \\
                               & 8  & 1.89m \\
                               & 16 & 3.79m \\
                               & 32 & 7.57m \\
        \bottomrule
    \end{tabular}
    \vspace{2mm}
    \caption{Trainable parameters, $T_p$, for various model sizes and Lora ranks.}
    \label{tab:decomposed_params}
\end{table}

\begin{figure}[t!]
    \centering
    \includegraphics[width=\linewidth]{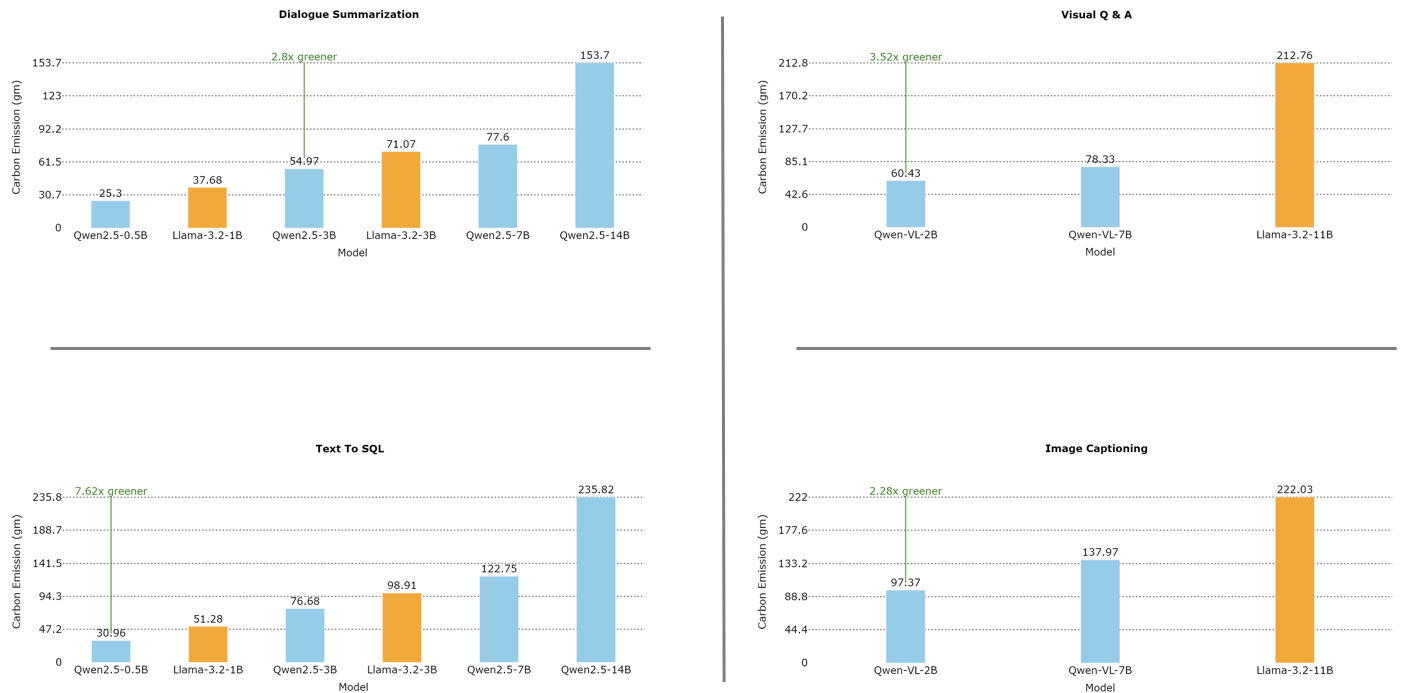}
    \caption{Comparison of Carbon Emissions Across Model Tasks. Bars with the green notation indicate the carbon efficiency of the best-performing fine-tuned model for each task, showing their efficiency in terms of carbon emissions relative to the highest carbon-emitting model for that task.}
    \label{fig:greener_model}
\end{figure}

For Text-to-SQL task, the Qwen2.5-0.5B model emits only 30.96 g of carbon, making it 7.62x greener(Figure \ref{fig:greener_model}) than the Qwen2.5-14B model, which has the highest emissions at 235.82 g. Similarly, in Image Captioning, the Qwen-VL-2B model is 2.28x greener than Llama-3.2-11B, showing the efficiency of smaller vision-language models in reducing carbon impact.

Across tasks, smaller models like Qwen2.5-0.5B and Qwen-VL-2B consistently demonstrate superior efficiency, achieving competitive performance with significantly lower carbon emissions. For instance, in Visual QA, Qwen-VL-2B is 3.52x greener than Llama-3.2-11B, while in Dialogue Summarization, Qwen2.5-0.5B is 2.8x greener than Qwen2.5-14B. These results underscore the importance of adopting fine-tuned smaller models and optimizing configurations to balance performance and sustainability. The significant carbon savings offered by these efficient models highlight the potential for sustainable AI practices, particularly in computationally intensive applications, without compromising performance.

%\section{CEGI illustration}

%Below illustrates how can we calculate CEGI index. For example, we performed few experiments with Qwen-VL-7B model for a solar panel defect classification dataset. The base model has F1 score 47.19 \% while the fine-tuning boosted the results to 78.65\%. In this case, $G_{M, \mu} (\%)$ can be calculated as below:

%\begin{equation}
%    \text{G}_{M,\mu} (\%) = \left( \frac{\mu_{T=\{F_T\}} - \mu_{R=\{B_M\}}}{\mu_{R=\{B_M\}}} \right) \times 100\% = \frac{0.075-0.001}{0.001} \times 100 = 7400
%\end{equation}
%If we assume that we are performing the experiment for single LoRA parameter with only one quantization, then we can take $|L_r| = 32$. The double summation operation in equation \ref{eq:gain_per_param_expanded} reduces to single scaler value in both numerator and denominator. The number of trainable parameters $\sum\limits_{\forall L_r} T_p = 3.123$ Millions. Hence, the Carbon Efficient Gain Index (CEGI) can be calculated as     $\frac{78.33 \times 1}{7400 \times 3.123} = 0.0033894$
\end{document}